\documentclass[final]{cvpr}

\usepackage{times}
\usepackage{epsfig}
\usepackage{graphicx}
\usepackage{subcaption}
\usepackage{amsmath}
\usepackage{amssymb}

\usepackage{bm}
\usepackage[ruled,vlined,linesnumbered]{algorithm2e}
\usepackage{comment}
\usepackage{color}
\usepackage{overpic}

\usepackage[pagebackref=true,breaklinks=true,colorlinks,bookmarks=false]{hyperref}

\def\versionArxiv{0}
\def\versionFinal{1}
\def\versionSupp{2}

\def\cvprPaperID{5298} 
\def\confYear{CVPR 2021}
\def\emitversion{\versionArxiv} 
\def\authorhighlights{0}

\if\emitversion\versionArxiv
\else
\fi

\makeatletter
\def\blfootnote{\gdef\@thefnmark{}\@footnotetext}
\makeatother

\newcommand{\myrefsec}[1]{Section~\ref{sec:#1}}
\newcommand{\myrefsecshort}[1]{(Sec.~\ref{sec:#1})}
\if\emitversion\versionFinal
    \newcommand{\myrefapp}[1]{\begin{NoHyper}Appendix~\ref{app:#1}\end{NoHyper}}
    
\else
    \newcommand{\myrefapp}[1]{Appendix~\ref{app:#1}}
    
\fi
\newcommand{\myreffig}[1]{Figure~\ref{fig:#1}}

\newcommand{\myrefFig}[1]{Figure~\ref{fig:#1}}
\newcommand{\myreffigTwo}[2]{Figures~\ref{fig:#1} and~\ref{fig:#2}}
\newcommand{\myrefalg}[1]{Algorithm~\ref{alg:#1}}

\newcommand{\myrefAlg}[1]{Algorithm~\ref{alg:#1}}
\newcommand{\myrefeq}[1]{Equation~(\ref{eq:#1})}
\newcommand{\myrefeqshort}[1]{(Eq.~(\ref{eq:#1}))}
\newcommand{\myreftab}[1]{Table~\ref{tab:#1}}

\newcommand{\mycite}[1]{\cite{#1}}
\newcommand{\myquote}[1]{``#1''}

\newcommand{\Expc}[2]{\mathbb{E}_{#1} #2}
\newcommand{\ExpcB}[2]{\mathbb{E}_{#1}\left[ #2 \right]}

\newcommand{\render}{\mathcal{R}}
\newcommand{\warp}{\mathcal{A}}
\newcommand{\disc}{\mathcal{D}}

\newcommand{\loss}{\mathcal{L}}

\newcommand{\degt}[1]{$#1^{\circ}$}

\newcommand{\rand}{\textnormal{rand}}
\newcommand{\genfield}{s}
\newcommand{\vel}{\mathrm{\mathbf{u}}}
\newcommand{\dens}{\rho}
\newcommand{\hull}{\bm{H}}
\newcommand{\light}{\bm{L}}

\newcommand{\image}{\bm{I}}
\newcommand{\imageRand}{\image_{\rand}}
\newcommand{\rendered}{\hat{\image}}
\newcommand{\near}{\mathrm{n}}
\newcommand{\far}{\mathrm{f}}
\newcommand{\nframes}{\bm{F}}
\newcommand{\iframes}{t}
\newcommand{\ffirst}{0}
\newcommand{\fprev}{\iframes-1}
\newcommand{\fcurr}{\iframes}
\newcommand{\fnext}{\iframes+1}
\newcommand{\flast}{\nframes-1}

\newcommand{\gstep}[2]{#1^{#2}}
\newcommand{\dfirst}{\gstep{\dens}{\ffirst}}
\newcommand{\dprev}{\gstep{\dens}{\fprev}}
\newcommand{\dcurr}{\gstep{\dens}{\fcurr}}
\newcommand{\dnext}{\gstep{\dens}{\fnext}}
\newcommand{\dlast}{\gstep{\dens}{\flast}}
\newcommand{\vfirst}{\gstep{\vel}{\ffirst}}
\newcommand{\vprev}{\gstep{\vel}{\fprev}}
\newcommand{\vcurr}{\gstep{\vel}{\fcurr}}

\newcommand{\nviews}{\bm{C}}
\newcommand{\iviews}{c}
\newcommand{\discTemp}{\disc_{\iframes}}

\newcommand{\shortGobWarp}{Glob-Trans}
\newcommand{\varSingle}{Single}
\newcommand{\varForward}{Forward}
\newcommand{\varCoupled}{C}
\newcommand{\varCoupledMS}{\varCoupled-MS}
\newcommand{\varGlobWarp}{\shortGobWarp}
\newcommand{\varDisc}{Full}
\newcommand{\varDiscTemp}{\varDisc-$\discTemp$}
\newcommand{\dig}{\tilde{\dens}}
\newcommand{\digPrev}{\gstep{\dig}{\fprev}}
\newcommand{\digCurr}{\gstep{\dig}{\fcurr}}
\newcommand{\digNext}{\gstep{\dig}{\fnext}}

\newcommand{\pder}[2]{\frac{\partial #1}{\partial #2}}
\newcommand{\pderInl}[2]{\partial #1 / \partial #2}

\newcommand{\target}{\textnormal{tar}}
\newcommand{\divergence}{\textnormal{div}}
\newcommand{\Ltarget}{\loss_{\target}}

\newcommand{\Lswarp}{\loss_{\warp(\genfield)}}
\newcommand{\Ldwarp}{\loss_{\warp(\dens)}}
\newcommand{\Lvwarp}{\loss_{\warp(\vel)}}
\newcommand{\Ldens}{\loss_{\dens}}
\newcommand{\Lvel}{\loss_{\vel}}

\newcommand{\Ldisc}{\loss_{\disc}}

\newcommand{\Ldenswarp}{\loss_{\warp(\dens)}}
\newcommand{\Lvelwarp}{\loss_{\warp(\vel)}}
\newcommand{\Ldiv}{\loss_{\divergence}}

\definecolor{picCol}{rgb}{1.0, 0.5, 0}
\newcommand{\mygraphicsT}[3]{%
\begin{overpic}[#1]{#2}%
    \put(2,93){\begin{scriptsize}{\color{picCol}#3}\end{scriptsize}}%
\end{overpic}%
}
\newcommand{\mygraphicsTtiny}[3]{%
\begin{overpic}[#1]{#2}%
    \put(2,93){\begin{tiny}{\color{picCol}#3}\end{tiny}}%
\end{overpic}%
}

\newcommand{\mygraphicsB}[3]{%
\begin{overpic}[#1]{#2}%
    \put(2,2){\begin{scriptsize}{\color{picCol}#3}\end{scriptsize}}%
\end{overpic}%
}

\definecolor{nCol}{rgb}{0.8, 0.3, 0}
\definecolor{eCol}{rgb}{0.2, 0.2, 0.8}
\definecolor{bCol}{rgb}{0.0, 0.6, 0.4}
\definecolor{notebarbaraCol}{rgb}{0.0, 0.8, 0.0}

\newcommand{\nt}[1]{}
\if\authorhighlights1
    \newcommand{\ef}[1]{{\color{eCol}#1}} 
\else
    \newcommand{\ef}[1]{#1} 

\fi

\newcommand{\myParSpace}{\vspace{-3pt}}

\SetKwProg{Fn}{Function}{:}{}%
\SetKwFunction{FnMax}{max}%
\SetKwFunction{FnLight}{BuildLighting}%
\SetKwFunction{FnResizeGrid}{ResizeGrid}%
\SetKwFunction{FnOptDensStep}{UpdateDensity}%
\SetKwFunction{FnOptVelStep}{UpdateVelocity}%
\SetKwFunction{FnOptDiscStep}{UpdateDiscriminator}%
\SetKwFunction{FnOptDens}{OptimizeDensity}%
\SetKwFunction{FnOptVel}{OptimizeVelocity}%
\SetKwFunction{FnOptFwd}{ReconstructForward}%
\SetKwFunction{FnOptMain}{ReconstructCoupled}%
\SetKwFunction{FnOptGwarp}{ReconstructGlobT}%
\newcommand{\algAssign}{\leftarrow}
\newcommand{\VarDiscActive}{useDisciminator}

\if\emitversion\versionFinal
\fi
\if\emitversion\versionSupp
\fi

\begin{document}

\title{Global Transport for Fluid Reconstruction with Learned Self-Supervision}

\author{Erik Franz\\
Technical University of Munich\\
{\tt\small erik.franz@tum.de}
\and
Barbara Solenthaler\\
ETH Zurich\\
{\tt\small solenthaler@inf.ethz.ch}
\and
Nils Thuerey\\
Technical University of Munich\\
{\tt\small nils.thuerey@tum.de}
}


\twocolumn[{%
	\renewcommand\twocolumn[1][]{#1}%
	\maketitle
	\begin{center}
		\vspace{-0.5cm}
    \includegraphics[width=\linewidth]{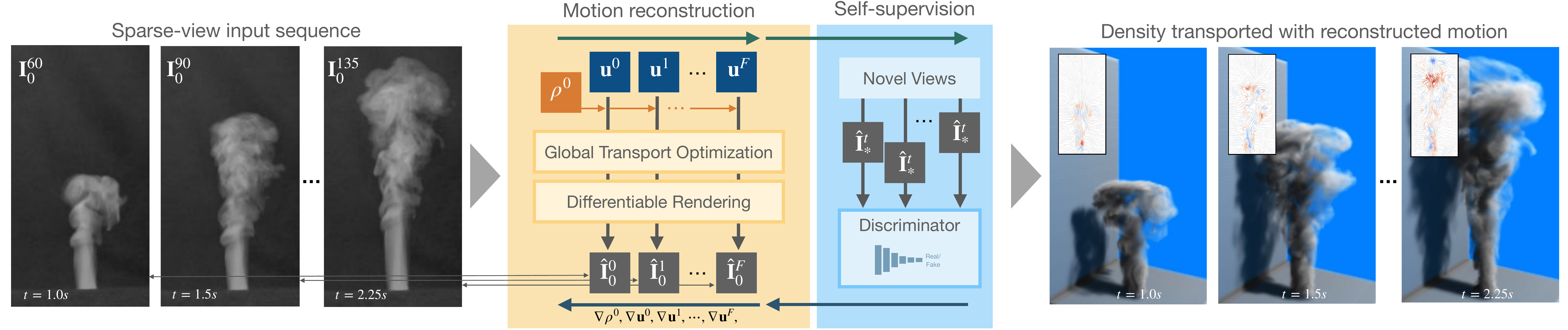}
    \captionof{figure}{
    We propose a novel algorithm that reconstructs the motion $\vel^{0..F}$ of a single initial state $\dens^0$ over the full course of $F$ frames of an input sequence, \ie, its {\em global transport}. Based on a learned self-supervision, our algorithm yields a realistic motion for highly under-constrained scenarios such as a single input view.
    }%
    \label{fig:intro.teaser}%
	\end{center}
}]

\begin{abstract}
We propose a novel method to reconstruct volumetric flows from sparse views via a global transport formulation. Instead of obtaining the space-time function of the observations, we reconstruct its motion based on a single initial state. In addition we introduce a learned self-supervision that constrains observations from unseen angles. These visual constraints are coupled via the transport constraints and a differentiable rendering step to arrive at a robust end-to-end reconstruction algorithm. This makes the reconstruction of highly realistic flow motions possible, even from only a single input view. We show with a variety of synthetic and real flows that the proposed global reconstruction of the transport process yields an improved reconstruction of the fluid motion.
\end{abstract}
\blfootnote{Source code: \url{https://github.com/tum-pbs/Global-Flow-Transport}}

\vspace{-0.2cm}

\section{Introduction}

The ambient space for many human activities and manufacturing processes is filled with fluids whose motion we can only observe indirectly \cite{morris2011dynamic,qian2017stereo,thapa2020dynamic,zangTomoFluidReconstructingDynamic2020}. Once passive markers, \eg, in the form of ink or dye, are injected into the fluid, it is possible to draw conclusions about the transport induced by the motion of the fluid. This form of motion reconstruction is highly important for a large variety of applications, from medical settings~\cite{peskin1972flow}, over engineering~\cite{shur1999detached} to visual effects~\cite{fedkiw:2001:VSO}, but at the same time poses huge challenges.

Fluids on human scales are typically turbulent, and exhibit a highly complex mixture of translating, shearing, and rotating motions \cite{pope2001turbulent}. In addition, the markers by construction need to be sparse or transparent in order not to fully occupy and occlude the observed volume. Several works have alleviated these challenges with specialized hardware \cite{Goldhahn07,Xiong:2017:rainbowPIV} or by incorporating the established physical model for fluids, the {\em Navier-Stokes} equations, into the reconstruction \cite{elsinga2006tomographic,Gregson:2014}. However, despite improvements in terms of quality of the reconstruction, the high non-linearity of the reconstruction problem coupled with ambiguous observations can cause the optimizations to find undesirable minimizers that deviate from ground truth motions.

In order to obtain a solution, existing approaches compute volumetric observations over time \cite{Gregson:2014,eckertScalarFlowLargescaleVolumetric2019,zangTomoFluidReconstructingDynamic2020}. 
Hence, despite \ef{including} a physical model, the observed quantities represent unknowns per time step, and are allowed to deviate from the constraints of the model. We make a central observation: when enforcing the physical model over the course of the complete trajectory of the observed fluid, the corresponding reconstructions 
yield a motion that better adheres to the ground truth. Thus, instead of a sequence, our reconstruction results in a {\em single initial state} of the density. 
This state is evolved via a {\em global transport}
over time purely by the physical model and the temporal reconstruction of the motion field. In addition to an improved reconstruction of the motion, this strict differentiable physics prior \cite{de2018end,hollLearningControlPDEs2019} allows us to work with very sparse observations, with only a single viewpoint in the extreme.


In order to better constrain the degrees of freedom in this single-viewpoint scenario, we  propose a method inspired by generative adversarial networks (GANs) \cite{goodfellow2014generative,xieTempoGANTemporallyCoherent2018}. Due to the complete lack of observations from other viewpoints, we make use of a small dataset of example motions, and train a convolutional neural network alongside the motion reconstruction that serves as a {\em discriminator}. This discriminator is evaluated for randomly sampled viewpoints in order to provide 
image space constraints that are coupled over time via our global transport formulation.

While existing works also typically focus on linear image formation  models~\cite{Ihrke:tomoFlames,Gregson:2012},
we combine the visual and transport constraints with a fully differentiable volumetric rendering pipeline. We account for complex lighting effects, such as absorption and self-shadowing, which are key elements to capture the visual appearance of many real-world marker observations. 

To summarize, the main contributions of our work are:
%
\vspace{-0.4em}
\begin{itemize}
  \itemsep0em
    \item A global multi-scale transport optimization via a differentiable physical model that yields purely transport-based reconstructions.
    \item A learned visual prior to obtain motions from very sparse and single input views.
    \item The inclusion of differentiable rendering with explicit lighting and volumetric self-shadowing.
\end{itemize}
\vspace{-0.2em}
To the best of our knowledge, these contributions make it possible for the first time
\ef{to construct a fluid motion from sparse views
in an end-to-end optimization, even from a single viewpoint}.
An overview is given in \myreffig{intro.teaser}.

\section{Related Work}

\paragraph{Flow Reconstruction}
Reconstructing fluid flows from observations has a long history in science and engineering. A wide variety of methodologies have been proposed, ranging from 
Schlieren imaging~\mycite{Dalziel00:BOS,Atcheson2008,Atcheson:2008:OEF},
over particle imaging velocimetry (PIV) methods~\mycite{Grant97:PIV,Elsinga:06},
to laser scanners~\mycite{Hawkins:timeVary,fuchs2007density} and structured light~\mycite{Gu:13} \ef{and light path~\mycite{Ji_2013_CVPR}} approaches.
Especially PIV has been widely used, and seen a variety of extensions and improvements, such as synthetic apertures~\mycite{belden2010three}, specialized lighting setups~\mycite{Xiong:2017:rainbowPIV}, and algorithms for custom hardware setups~\mycite{fahringer2015volumetric}.  
In the following, we focus on visible light capturing approaches, as they avoid the reliance on specialized, and often expensive hardware. 

This capturing modality is especially interesting in conjunction with
sparse reconstructions, \eg, medical tomography settings often favor view sparsity  \mycite{sidky2008image,chen2012thoracic,shen2019r}. 
Here, the highly under-constrained setting of a large number of degrees of freedom of a dense grid to be reconstructed from a very small number of input views is highly challenging for reconstruction algorithms.
One avenue to obtain a reduced solution space is to introduce physical priors. For fluids, the {\em Navier-Stokes equations}~\mycite{pope2001turbulent} represent a well-established physical model for incompressible fluids. 
\ef{
\Eg, the adjoint method was used for gradient-based optimization of volumetric flows, such as fluid control \mycite{McNamara2004}.
On the other hand, Gregson \etal \mycite{Gregson:2014} introduced a multi-view convex optimization algorithm that makes use of a discrete advection step and a divergence-free projection.
This approach was extended to incorporate view-interpolation to obtain additional constraints~\mycite{zangTomoFluidReconstructingDynamic2020}.
Convex optimization was also used for the reconstruction from sparse views \mycite{eckertScalarFlowLargescaleVolumetric2019}, 
or specialized single view reconstruction~\mycite{eckertCoupledFluidDensity2018}. However, this requires a hand-crafted regularizer \ef{to suppress} depth-aligned motions.
In contrast, we rely on gradients provided by deep-learning methods for our reconstructions.}
\ef{And w}hile several of these methods introduce transport terms for inter-frame constraints, none
introduces an \myquote{end-to-end} constraint that yields a single, physical transport process from an initial state to the last observed frame.
\ef{Such gradient propagation through PDEs
is also being explored in differentiable physical simulation \mycite{hollLearningControlPDEs2019}.}


\myParSpace{}
\myParSpace{}
\paragraph{3D Reconstruction and Appearance}
\ef{Solving the tomography for static scattering volumetric media
has been studied via inverse scattering \mycite{Gkioulekas2016}
or for large-scale cloud-scapes \mycite{Levis_2015_ICCV}.}
\ef{Although we target volumetric reconstruction,}
our goals \ef{also} partially align with settings where a clearly defined surface is visible, \eg, reconstructions of 3D geometry \mycite{musialski2013survey,koutsoudis2014multi}\ef{, including its deformation \mycite{Zang2018}}.
Mesh-based methods were proposed to create a deformation of a labeled template mesh \mycite{kanazawaLearningCategorySpecificMesh2018},
or of spherical prototypes \mycite{katoNeural3DMesh2018}. 
While marker density volumes do not exhibit a clear surface,
they share similarities with voxel-based reconstructions \mycite{papon2013voxel,moon2018v2v}. In addition, learned representations 
\mycite{sitzmann2019deepvoxels,lombardiNeuralVolumesLearning2019,sitzmann2019scene,mildenhall2020nerf} have the flexibility to encode information about transparent materials and their reflectance properties, and temporally changing deformations \mycite{mescheder2019occnet,niemeyer2019occupancy}.

\ef{In conjunction with sparse-view reconstructions, regularization becomes increasingly important to obtain a meaningful solution.}
One line of work has employed different forms of appearance transfer \mycite{efros2001image,javed2005appearance}, \eg, 
to match the histograms of fluid reconstructions \mycite{okabeFluidVolumeModeling2015}, while others have proposed view interpolations via optical flow 
\mycite{zangTomoFluidReconstructingDynamic2020}.
In the context of style transfer for natural images, learned approaches via GANs \mycite{goodfellow2014generative,RadfordMC15} were shown to be especially powerful \mycite{zhu2017unpaired,karras2019style}.
GANs were likewise employed in fluid synthesis settings \mycite{xieTempoGANTemporallyCoherent2018}, and we propose a learned discriminator that works alongside a single flow reconstruction process.

A similar idea was used for single-view tomographic reconstruction of static objects: The approach by Henzler \etal \mycite{henzlerSingleimageTomography3D2018} was extended to include a learned discriminator for a specific class of objects \mycite{henzlerEscapingPlatoCave2019} to constrain unseen views.
We extend this approach to sequences, and show that the concept of learned self-supervision provides a highly useful building block for physical reconstructions.

\myParSpace{}
\myParSpace{}
\paragraph{Differentiable Rendering}
Computing derivatives of pixels in an image with respect to input variables is essential for many inverse problems. 
Several fast but approximate differentiable rendering methods have been presented that model simple light transport effects \mycite{Loper2014,kato2018neural}.
Physics-based neural renderers account for secondary effects, such as shadows and indirect light. Li \etal \mycite{Li2018DMC} presented the first general-purpose differentiable ray tracer, and Nimier \etal \mycite{Nimier2019Mitsuba2} a versatile MCMC renderer that was, among other examples, applied to smoke densities.
Neural renderers \ef{\mycite{tewari2020state}} were used for 3D reconstruction problems, \eg, to render an explicit scene representation  \cite{henzlerEscapingPlatoCave2019} or to synthesize novel views \cite{mildenhall2020nerf}.

In the context of fluid simulations, differentiable renderers were used with and without differentiable solvers to optimize robotic controllers~\mycite{Schenk2018}, to initialize a water wave simulator~\mycite{Hu2020}, and to transfer a style from an input image onto a 3D density field~\mycite{Kim2019,Kim2020}.
Additive, \ie, linear, lighting models were often used for flow reconstruction~\mycite{eckertScalarFlowLargescaleVolumetric2019, zangTomoFluidReconstructingDynamic2020}. They are suitable for thin marker densities that have little absorption under uniform lighting conditions.
In contrast, our method employs a differentiable rendering model that handles non-linear attenuation and self-shadowing.

\section{Method}

The central goal of our algorithm is to reconstruct a space-time sequence of volumetric fluid motions $\vel$ such that a passively transported
field of marker density $\dens$ matches a set of target images. These targets are given as $\iviews \in \nviews$ calibrated input views $\image^{\iframes}_{\iviews}$ for a time sequence of $\nframes$ frames with $\{\iframes|\iframes \in \mathbb{N}_0 , \iframes < \nframes\}$.
We represent $\dens$ and $\vel$ as dense Eulerian grids, which are directly optimized through iterative gradient updates calculated from the proposed loss functions and propagated through the full sequence to achieve a solution that adheres to a global transport formulation.

\subsection{Physical Prior} \label{sec:method.transport}

We derive our method on the basis of a strong physical prior given by the Navier-Stokes equations that describe the motion of 
an incompressible fluid:
\begin{equation}\label{eq:navier.stokes}
\begin{aligned}
  \vel_{t} + \vel \cdot \nabla \vel = -\nabla{p} + \nu\nabla^2\vel 
  \ , \text{ s.t. } \  
   \nabla\cdot\vel = 0. 
\end{aligned}
\end{equation}
In addition to the fluid velocity $\vel$, $p$ denotes the pressure and $\nu$ the kinematic viscosity.
\ef{We implement these equations via a differentiable transport operator $\warp(\genfield, \vel)$
that can advect a generic field $\genfield$ (can be $\dens$ or $\vel$) using a velocity field $\vel$.}
For the discretized operator $\warp$ we use a second-order transport scheme to preserve small-scale content on the transported fields \cite{selleUnconditionallyStableMacCormack2008}.
Details are in \myrefapp{impl.warp}.

\myParSpace{}%
\myParSpace{}%
\paragraph{Global Density Transport Optimization} \label{sec:method.transport.full}
\ef{Given an initial density $\dfirst$, the state $\dcurr$ at time $\iframes$ is given by}
\begin{equation}\label{eq:method.transport.warp}
    \dcurr := \gstep{\warp}{\iframes}(\dfirst) = \warp(\warp(\warp(\dfirst,\vfirst), \gstep{\vel}{1}) \dots ,\vprev)
\end{equation}
\ef{
and we only optimize $\dfirst$ and the velocity sequence $\{\vfirst,\dots,\vprev\}$, instead of optimizing all $\dcurr$ individually, as done by previous approaches.
With this formulation}
we enforce a correct transport over the entire fluid flow trajectory, resulting in a reconstructed motion that better adheres to \myrefeq{navier.stokes}.
\ef{For optimization, all $\dcurr$ are constrained by a \ef{composite} loss $\loss_{\dens}$, that we detail in \myrefsec{recon}. Thus,}
the gradient for $\dfirst$ is the sum of derivatives of
$\loss_{\dens}$, of all frames $\iframes \in \nframes$, \wrt to the density of the first frame:  
\begin{equation}\label{eq:method.transport.gradient}
    \nabla \dfirst_{\warp} = \sum_{\iframes=0}^{\nframes} \pder{\loss_{\dens}(\dcurr)}{\dfirst}
    = \sum_{\iframes=0}^{\nframes} \pder{\gstep{\warp}{\iframes}(\dfirst)}{\dfirst} \pder{\loss_{\dens}(\gstep{\warp}{\iframes}(\dfirst))}{\gstep{\warp}{\iframes}(\dfirst)}.
\end{equation}
Back-propagation of the gradients $\nabla \dens^{\fnext}$ through every transport step $\warp(\dens^t, \vel^t)$ takes the iterative form
\begin{equation}\label{eq:method.transport.backwarp}
\begin{aligned}
    \nabla \dcurr_{\warp} = \nabla \dcurr + \pder{\warp(\dcurr, \vcurr)}{\dcurr}\nabla \dnext_{\warp}
\end{aligned}
\end{equation}
where $\nabla \dens^{t} = \pderInl{\loss_{\dens}(\dcurr)}{\dcurr}$ are the individual per-frame gradients and $\nabla \dlast_{\warp} = \nabla \dlast$.
In practice, we found that using an exponential moving average (EMA) with decay $\beta$ to accumulate the gradients, \ie,
\begin{equation}\label{eq:method.transport.ema}
    \nabla \dcurr_{\warp} = (1-\beta)\nabla \dcurr + \beta \pder{\warp(\dcurr, \vcurr)}{\dcurr}\nabla \dnext_{\warp},
\end{equation}
leads to better defined structures and smoother motion.
This construction, which we will refer to as {\em global transport} formulation, 
requires an initial velocity sequence to create $\gstep{\warp}{\iframes}(\dfirst)$, which we acquire from a  forward-only pre-optimization pass, as described in  \myrefsec{recon}.
\ef{The velocity also receives gradients $\nabla \vcurr_{\warp} = \sum_{i=\iframes}^{\nframes} \pderInl{\Ldens(\dens^{i})}{\vcurr} = \pderInl{\warp(\dcurr, \vcurr)}{\vcurr}\nabla \dnext_{\warp}$ from the back-propagation through the density advection. Those are, however, not back-propagated further through velocity self-advection.}

\ef{
\myParSpace{}%
\myParSpace{}%
\paragraph{Transport Losses}
A per-frame} transport loss that takes the general form of 
\begin{equation}\label{eq:loss.generic.warp}
\begin{aligned}
\Lswarp=|\warp(\gstep{\genfield}{\fprev},\vprev) - \gstep{\genfield}{\fcurr}|^2 
 + |\warp(\gstep{\genfield}{\fcurr},\vcurr) - \gstep{\genfield}{\fnext}|^2
\end{aligned}
\end{equation}
connects $\genfield^{\fcurr}$ to the previous and subsequent frames at times $\fprev$ and $\fnext$.
\ef{The self-advection loss $\Lvelwarp$ induces temporal coherence in the velocity fields
while the density transport loss $\Ldwarp$ is used to build the initial velocity sequence.
Despite the inclusion of and back-propagation through the global transport via \myrefeq{method.transport.ema}, both still contribute important gradients that help to reconstruct the inflow region and the motion within the visual hull, 
which we show with an ablation in \myrefapp{eval.warploss}.}
%
The divergence-free constraint of \myrefeq{navier.stokes} is enforced by \ef{additionally} minimizing
\begin{equation}\label{eq:loss.vel.div}
\begin{aligned}
\Ldiv=\left| \nabla \cdot \vel \right|^2.
\end{aligned}
\end{equation}

\subsection{Differentiable Rendering}\label{sec:method.render}
\ef{To connect the physical priors to the target observations}
%
we leverage recent progress in GD-based differentiable rendering~\mycite{katoNeural3DMesh2018, henzlerEscapingPlatoCave2019}.
We employ a non-linear image formation (IF) model that includes attenuation and shadowing
to handle varying lighting conditions and denser material while still being able to solve the underlying tomographic reconstruction problem.
Our rendering operator $\render()$ creates the image seen from view $\iviews$ of the density $\dens$ given the outgoing light $\light$ at each point in the volume by solving the integral
\begin{equation}
\begin{aligned}
    \render(\dens,\light,\iviews) = \int_{\near}^{\far} \light(x) e^{-\int_{\near}^{x} \dens(a) d a} d x
\end{aligned}\label{eq:render}
\end{equation}
for each pixel-ray $\near \rightarrow \far$ of the image, using \ef{the} Beer-Lambert law for absorption \mycite{engel2004real}.

\ef{The outgoing light $\light$ is computed from
point lights $p \in P$ at positions $x_p$ with inverse-square falloff and single-scattering as well as}
ambient lighting to approximate the effect of multi-scattering. The intensities are $i_p$ and $i_a$, respectively.
The total outgoing light at point $x$ is given as
\begin{equation}
\begin{aligned}
\light_{\dens}(x) = i_a \dens(x) + \sum_{p\in P} i_p \dens(x) \frac{1}{1+||x_p-x||_2 } e^{-\int_{x_p}^x\dens(a) d a}.
\end{aligned}\label{eq:light}
\end{equation}

\ef{In practice we employ ray-marching
to approximate the shadowing term and render the image and its transparency.
Implementation details and
an evaluation of the rendering model
can be found in \myrefapp{impl.render}.
}

With this formulation we can constrain \ef{the renderings of the reconstruction to the target images $\gstep{\image}{\fcurr}_{\iviews}$ with}
\begin{equation}\label{eq:loss.dens.tar}
    \Ltarget=\frac{1}{c}\sum_{\iviews \in \nviews}|\gstep{\image}{\fcurr}_{\iviews} - \render(\dcurr,\light_{\dcurr},c)|^2.
\end{equation}

\begin{figure*}[t!]
    \centering
    \mygraphicsT{width=0.11\linewidth}{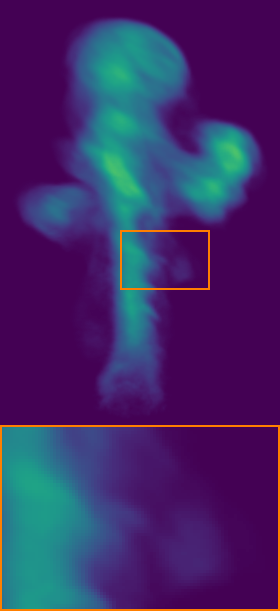}{a) \varSingle}%
    \mygraphicsT{width=0.11\linewidth}{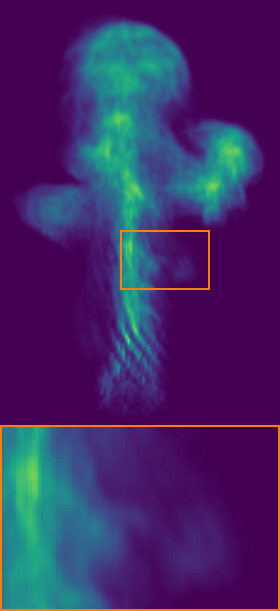}{b) \varForward}%
    \mygraphicsT{width=0.11\linewidth}{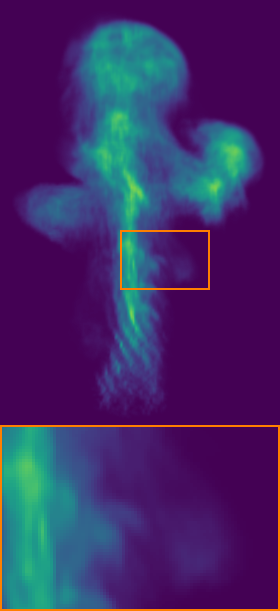}{c) \varCoupled}%
    \mygraphicsT{width=0.11\linewidth}{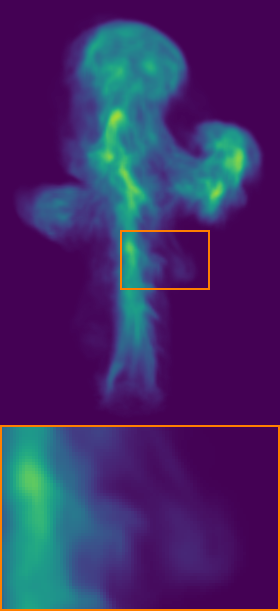}{d) \varCoupledMS}%
    \mygraphicsT{width=0.11\linewidth}{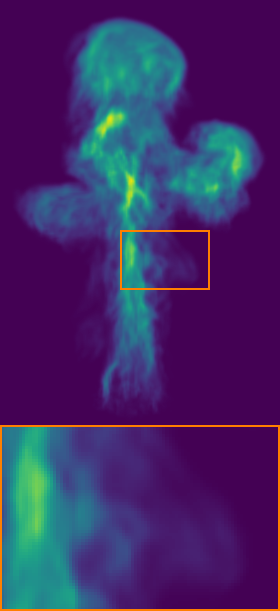}{e) \varGlobWarp}%
    \mygraphicsT{width=0.11\linewidth}{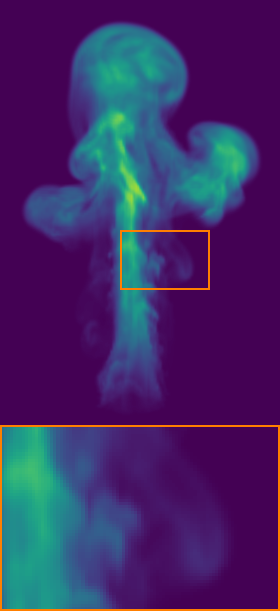}{f) Reference}%
    \hfill%
    \mygraphicsT{width=0.11\linewidth}{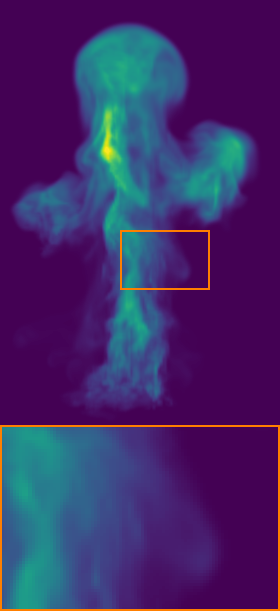}{g) SF \mycite{eckertScalarFlowLargescaleVolumetric2019}}%
    \mygraphicsT{width=0.11\linewidth}{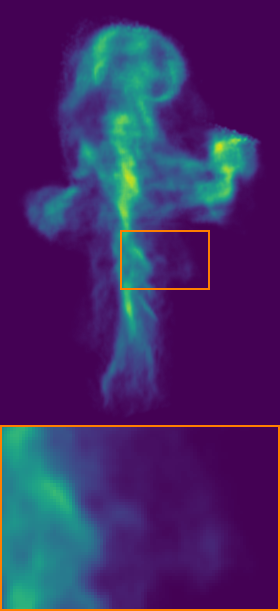}{h) TF \mycite{zangTomoFluidReconstructingDynamic2020}}%
    \mygraphicsT{width=0.11\linewidth}{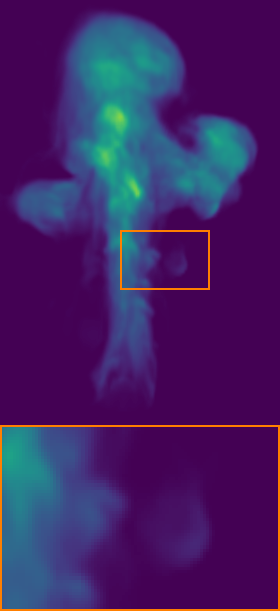}{i) NV \mycite{lombardiNeuralVolumesLearning2019}}%
    \caption{
    Multi-view evaluation with synthetic data: (a-e) Ablation with reference shown in (f). The different versions of the ablation (details in \myrefsec{res.multi}) continually improve density reconstruction and motion. 
    (g-i) Comparison with previous work 
    ScalarFlow \mycite{eckertScalarFlowLargescaleVolumetric2019}, TomoFluid \mycite{zangTomoFluidReconstructingDynamic2020}, and NeuralVolumes \mycite{lombardiNeuralVolumesLearning2019}. Our method in (e) yields an improved density reconstruction, in addition to a coherent and physical transport (\ef{see also} \myreftab{eval.synth}). 
    }
    \label{fig:eval.synth}
\end{figure*}

\subsection{Learned Self-Supervision} 
\label{sec:method.disc}
Our physical priors \ef{and rendering} constraints enable the reconstruction of a \ef{physically correct} flow even from sparse views.
However, $\dens$, and therefore the transport,
\ef{are increasingly under-constrained with fewer views,}
and the huge space of indistinguishable solutions can lead to
a mismatch between real and computed motions\ef{, severely reducing accuracy when using only a single view}.
To address these ambiguities, we train a fully convolutional neural network alongside the motion reconstruction
that serves as a discriminator between images of real and reconstructed smoke.
\ef{This} discriminator learns to transfer features of real flows to the reconstruction via their visual projections over the course of the optimization.
\ef{While the discriminator can not recover the true volumetric structure pertaining to a given sample, it constrains the results to be plausible \wrt the given target samples.}
We chose the RaLSGAN (relativistic average least squares) loss variant~\mycite{jolicoeur-martineauRelativisticDiscriminatorKey2018a} as our discriminator loss for increased training stability:
\begin{equation}
\begin{aligned}
    \Ldisc(\dens,l) := &\ExpcB{r \sim R}{ \left( \disc(\image_r) - \Expc{f \sim \Omega}{ \disc(\rendered_f)} -l \right)^2}\\
    + &\ExpcB{f \sim \Omega}{\left(\disc(\rendered_f) - \Expc{r \sim R}{\disc(\image_r)} +l \right)^2}.
\end{aligned}
\label{eq:loss.disc}
\end{equation}
$\image_r$ are randomly sampled from a set of reference images $R$,
and $\rendered_f := \render(\dens, \light_{\dens}, f)$ are rendered random views of the reconstruction.
The label $l$ is $1$ when training the discriminator, and $-1$ when using it as a loss to optimize the density via $\pderInl{\Ldisc(\dens,-1)}{\dens}$.
This auxiliary loss term provides gradients for the optimization process and in this way constrains the large space of possible solutions for the fluid motion. Further details are given in \myrefapp{impl.disc}.

Extending the discriminator in the temporal domain via a triplet of inputs ($\image^{t-i}, \image^{t}, \image^{t+i}$) was proposed for fluid in other contexts \mycite{xieTempoGANTemporallyCoherent2018}. We show in \myrefsec{eval.1view} that such an extension does not yield benefits as our transport priors already successfully constrain the physicality of the motion.


\subsection{\ef{Additional Constraints}}\label{sec:method.other}
During reconstruction we restrict the density to the visual hull $\hull$ constructed from the target views $\nviews$. After background subtraction the targets are turned into a binary mask,
and projected into the volume ($\render^{-1})$ to create the hull via intersection:
\begin{equation} \label{eq:hull}
    \hull^t := \bigcap_{\iviews \in \nviews} \render^{-1}( H( \gstep{\image}{\fcurr}_{\iviews} - \epsilon), \iviews) \ ,
\end{equation}
with $H$ being the Heaviside step function.
Using the visual hull as hard constraint avoids residual density and tomography artifacts in outer regions of the reconstruction volume.
For single-view reconstruction the hull is constructed using auxiliary image hulls created by rotation of the original view around a symmetry axis of the physical setup \ef{(\myrefapp{sup.hull}).
While this hull is itself not physically correct, it constraints the solution to plausible shapes.}

As we focus on fluid convection scenarios such as rising smoke plumes, we model dedicated inflow regions 
\ef{for $\dens$ that are degrees of freedom, optimized together with the flow.}
We allow for free motions via Neumann boundary conditions for $\vel$.

To reduce the non-linearity of the transport constraints, we follow a commonly employed multi-scale approach~\cite{meinhardt2013horn,Gregson:2014} that first optimizes larger structures without being affected by local gradients of small-scale features. This is especially important for the global formulation, as it enables the optimization of large-scale displacements when constructing the velocity field, and makes it easier to handle larger features in the density content, without details obstructing the velocity transport gradients,
\ef{as} $\pderInl{\warp(\genfield,\vel)}{\vel}$ depends on the spatial gradients of $\genfield$ \ef{(see also \myrefapp{impl.warp})}.

\section{Reconstruction Framework}\label{sec:recon}

To reconstruct a specific flow, the density $\dens$ and velocity $\vel$ are first initialized  with random values inside the visual hull $\hull$.
We alternately update $\dens$ and $\vel$ via Adam iterations \cite{kingma2014adam}.
The targets $\gstep{\image}{\fcurr}_{\iviews}$ for the rendering loss are given as image sequences with their corresponding camera calibrations $\nviews$. 
When using targets with a background, we composite the background and the reconstruction using the accumulated \ef{attenuation} of the volume in order to match the appearance of the targets.
\ef{While we experimented with optimizing the lighting (\myrefapp{sup.bkg-lightopt}), we use light positions and intensities that were determined empirically in our reconstructions.}
Below we give an overview of our framework, while algorithmic details and parameter settings are given in \ef{\myrefapp{sup.alg}}.


\myParSpace{}
\myParSpace{}
\paragraph{Pre-Optimization Pass}
We run a single \textit{forward} pass to obtain a density sequence $\dig$ as initial guess, and to compute an initial velocity sequence as starting point for the global transport optimization from \myrefsec{method.transport.full}. 
The density is computed via tomography using $\Ltarget$ and the visual hull $\hull$ without transport constraints.
The velocities $\vcurr$ are optimized using $\digCurr$ and $\digNext$ via the density transport $\Ldwarp$ and $\Ldiv$. 
To improve temporal coherence in this step the frames are initialized from the transported previous frame before optimization: 
$\digCurr:=\warp(\digPrev, \warp(\gstep{\vel}{\fcurr -2})), \vcurr:=\warp(\vprev,\vprev)$,
for which the initial velocity $\vfirst$ is constructed using the multi-scale scheme of \myrefsec{method.transport}.
The initial density sequence $\dig$ is discarded when the global transport constraints are introduced.

\myParSpace{}
\myParSpace{}
\paragraph{Coupled Reconstruction}
The main part of the reconstruction couples the observations over time by incorporating the remaining priors, namely $\Ldwarp$ for the density and $\Lvwarp$ for velocity. 
Combined, for the density loss in \myrefeq{method.transport.gradient} this  yields $\loss_{\dens} = \Ltarget + \Ldwarp + \Ldisc(\dens,-1)$.
%
Subsequent iterations of the optimizer run over the whole sequence, updating degrees of freedom for all time steps.
When optimizing via the global transport formulation
\ef{\myrefeqshort{method.transport.ema}}
the degrees of freedom for densities are reduced to $\dfirst$, and a density sequence is created by advecting $\dfirst$ forward before updating the velocities via back-propagation through the sequence while accumulating the gradients for $\dfirst$.
The multi-scale approach is realized by synchronously increasing the resolution for all densities and velocities of the complete sequence with an exponential growth factor of $\eta=1.2$ in fixed intervals.
Target and rendering resolution for $\Ltarget$ and $\disc$ are adjusted in accordance with the multi-scale resolutions.


\section{Results and Evaluation}\label{sec:res}

We use a synthetic data set with known ground truth motion 
to evaluate the performance for multiple target views, before evaluating real data for the single-view case.
For all cases below, please see the supplemental video for details. The differences of the velocity reconstructions are especially apparent when seen in motion.
\ef{The video and source code can be found on the project website.}

\subsection{Multi-view Reconstructions}\label{sec:res.multi}
We simulate a fluid flow with a publicly available Navier-Stokes solver~\cite{mantaflow} with a resolution of $128 \times 196 \times 128$. Observations are generated for 120 frames over time with 5 viewpoints spread over a 120 degree arc using real calibrations. 

\myParSpace{}
\myParSpace{}
\paragraph{Ablation}
We first illustrate our algorithm with an ablation study.
A qualitative comparison of these versions is shown in \myreffig{eval.synth}.
The simplest \textit{single} version uses the single pass described in \myrefsec{recon}.
It yields a very good reconstruction of the input views, and low image space and volumetric errors for $\dens^{\hull}$ and $\imageRand$ in \myreftab{eval}, respectively, but strong tomographic artifacts.
Due to the known deficiences of direct metrics like RMSE for complex signals, we additionally compute SSIM~\mycite{wang2004ssim} and LPIPS~\mycite{zhang2018lpips} metrics for $\imageRand$.
Although the single-pass version has good SSIM and LPIPS scores,
it results in incoherent and unphysical motions over time. This is visible in 
terms of the metric $\warp(\dens)^{\hull}$ in \myreftab{eval}, which measures how well the 
motion from one frame explains the next state, \ie, $\warp(\dens)^{\hull}=|\warp(\dens^{t-1}) - \dens^t|_2$.
%

\begin{table*}
    \centering
    \begin{subtable}[t]{0.65\textwidth}
        \centering
        \begin{footnotesize}
        \begin{tabular}{|l|c|c|c|c|c|c|c|}\hline
            \multicolumn{1}{|r|}{Metrics} & $\warp(\dens)^{\hull}$ & \multicolumn{2}{c|}{$\dens^{\hull}$} &  \multicolumn{2}{c|}{$\vel^{\hull}$} & \multicolumn{2}{c|}{$\imageRand$} \\\hline
           \multicolumn{1}{|l|}{Method} & RMSE$\downarrow$ & RMSE$\downarrow$ & SSIM$\uparrow$ & RMSE$\downarrow$ & SSIM$\uparrow$  & PSNR$\uparrow$ & LPIPS$\downarrow$\\\hline\hline
            \varSingle 
                & 0.1352 & 1.300 & 0.599 & 0.453 & 0.148 & 41.51 & .0203 \\\hline
            \varForward 
                & 0.1222 & 2.429 & 0.464 & 0.463 & 0.165 & 39.81 & .0281 \\\hline
            \varCoupled 
                & 0.0291 & 2.268 & 0.502 & 0.452 & 0.180 & 40.26 & .0244 \\\hline
            \varCoupledMS 
                & 0.0505 & 1.309 & 0.585 & 0.445 & 0.167 & 41.59 & .0186 \\\hline
            \hline%
            \hline%
            \varGlobWarp 
                & \textbf{1.4e-7} & 1.844 & \textbf{0.533} & 0.452 & 0.171 & \textbf{39.86} & \textbf{.0257} \\\hline
            SF~\mycite{eckertScalarFlowLargescaleVolumetric2019} 
                & 1.573* & 2.814 & 0.328 & \textbf{0.440} & \textbf{0.220} & 35.80 & .0529 \\\hline
            TF~\mycite{zangTomoFluidReconstructingDynamic2020} 
                & 0.3879 & \textbf{1.707} & 0.473 & 0.544 & 0.082 & 30.19 & .0492 \\\hline
            NV~\mycite{lombardiNeuralVolumesLearning2019} 
                & N/A & 2.002 & 0.337& N/A & N/A & 27.77 & .0674 \\\hline
        \end{tabular}
        \end{footnotesize}
        \caption{Multi-view: mean volume and image statistics and errors for  synthetic data.
        }
        \label{tab:eval.synth}
    \end{subtable}%
    \hfill%
    \begin{subtable}[t]{0.34\textwidth}
        \centering%
        \begin{footnotesize}%
        \begin{tabular}{|l|c|c|c|c|}\hline%
            \multicolumn{1}{|r|}{Metrics}  & \multicolumn{2}{c|}{Target views} & $\imageRand$ \\\hline%
            Method & PSNR$\uparrow$ & LPIPS$\downarrow$ & FID$\downarrow$\\\hline\hline%
            \varForward 
                & 29.42 & .0538 & 164.6 \\\hline%
            \varCoupledMS 
                & 28.31 & .0541 & 166.7 \\\hline%
            \varGlobWarp 
                & 27.29 & .0484 & 151.2 \\\hline%
            \hline%
            \hline%
            \varDisc
                & 27.40 & .0489 & \textbf{140.3} \\\hline%
            \varDisc$+\disc_{\iframes}$ 
                & \textbf{27.49} & \textbf{.0478} & 147.3\\\hline%
            pGAN~\mycite{henzlerEscapingPlatoCave2019} 
                & 22.91 & .0891 & 185.4\\\hline%
        \end{tabular}%
        \end{footnotesize}%
        \caption{Single-view: mean image statistics evaluated for five new random views.
        }%
        \label{tab:eval.real}%
    \end{subtable}
    \caption{
    Error metrics for multi-view (left) and single-view cases (right). 
    $\warp(\dens)^{\hull}$ denotes the motion difference $\warp(\dens^{t-1}) - \dens^t$, 
    $\dens^{\hull}$ and $\vel^{\hull}$ denote volumetric errors; 
    all of these are measured inside the visual hull.
    $\imageRand$ 
    measures differences to rendered targets using 32 random views.
    Bold numbers denote best scores for full algorithms, \ie, without ablation variants.
    The * for $\warp(\dens)^{\hull}$ of SF indicates resampling the velocities to the same grid resolution as other methods for comparison.
    } \label{tab:eval}
    \vspace{-0.2cm}
\end{table*}

The \textit{forward} reconstruction (\myrefsec{recon}) introduces a temporal coupling, 
and yields slight improvements in terms of temporal metrics and reduces the artifacts.
Next, enabling \textit{coupling} (version \varCoupled) over time by optimizing the sequence with transport losses 
greatly improves the transport accuracy to approx. 25\% of the error of the forward version. 
Version \varCoupledMS\ in \myreftab{eval} activates
\textit{multi-scale} reconstruction during the coupled optimization and yields an improved reconstruction of larger structures, at the expense of an increased transport error.
With version \varGlobWarp~we arrive at our \textit{global transport} optimization, as described in \myrefsec{method.transport.full}.
It results in sharper reconstructions, matching the perceived sharpness of the reference.
This version achieves very good reconstructions in terms of density errors, and ensures a correct transport in terms of $\warp(\dens)^{\hull}$ up to numerical precision, \ef{see} \myreftab{eval}.

\begin{figure}%
    \mygraphicsB{height=0.424\linewidth}{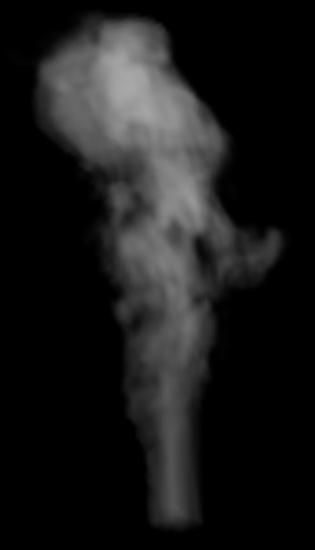}{a) NV \mycite{lombardiNeuralVolumesLearning2019}}%
    \hfill%
    \mygraphicsB{height=0.424\linewidth}{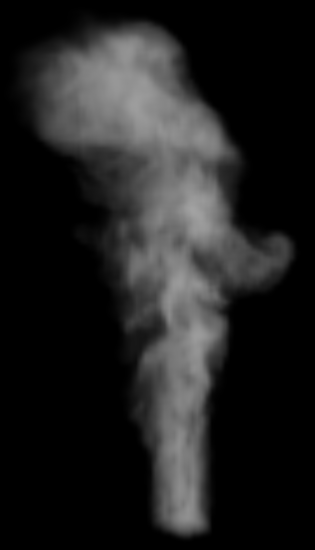}{b) SF \mycite{eckertScalarFlowLargescaleVolumetric2019} }%
    \hfill%
    \mygraphicsB{height=0.424\linewidth}{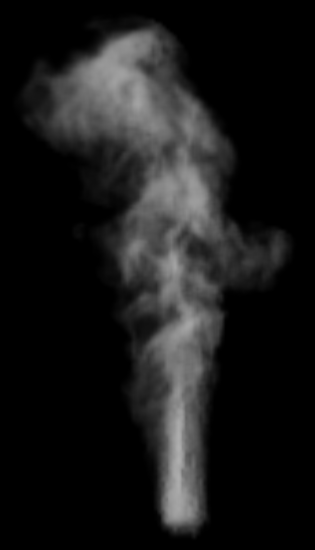}%
    {\shortstack[l]{c) \varGlobWarp}}%
    \hfill%
    \mygraphicsB{height=0.424\linewidth}{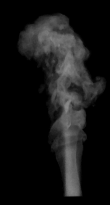}{\shortstack[l]{d) Nearest Target}}%
    \caption{
    Other methods for multi-view reconstruction (a,b)
    compared to our approach (c), versus the nearest target from 
    a \degt{30} rotated viewpoint (d). Even for the new viewpoint, our reconstruction matches the structures of the target, while the other methods produce overly smooth reconstructions.
    The SF~\mycite{eckertScalarFlowLargescaleVolumetric2019} result is representative for  TF~\mycite{zangTomoFluidReconstructingDynamic2020} here.
    }%
    \label{fig:eval.realTar.5view}%
\end{figure}%

\myParSpace{}
\myParSpace{}
\paragraph{State of the Art}\label{sec:eval.soa}

We compare our global transport approach to three state of the art methods.
Specifically, we compare with
{\em ScalarFlow} (SF)~\mycite{eckertScalarFlowLargescaleVolumetric2019} and
{\em TomoFluid} (TF)~\mycite{zangTomoFluidReconstructingDynamic2020}  
as representatives of physics-based reconstruction methods, and {\em NeuralVolumes} (NV)~\cite{lombardiNeuralVolumesLearning2019} as a purely visual, learning-based method.
We use the synthetic data set from before, while targets are rendered in accordance with each method for fairness.
Quantitative results are summarized in \myreftab{eval}, while qualitative comparisons are depicted in \myreffig{eval.synth} (g-i). 

The SF reconstruction \mycite{eckertScalarFlowLargescaleVolumetric2019} employs physical priors in a single forward pass, similar to a constrained physics-based simulation.
This leads to a temporally coherent velocity, but underperforms in terms of matching the target views. 
Note that SF has a natural advantage for the velocity reconstruction metrics in \myreftab{eval} for $\vel^{\hull}$ as it is based on the same solver used to produce the synthetic data (see \myrefapp{eval}). 
The TF method \cite{zangTomoFluidReconstructingDynamic2020} improves upon this behavior by employing an optimization with physical priors, similar to the {\em coupled} variant of our ablation. However, due to its reliance on inter-frame motions and spatio-temporal smoothness, the computed motion is temporally incoherent. It additionally relies on a view-interpolation scheme to constrain intermediate view directions,
which can be severely degraded far from the target views, \eg, as shown in \myreffig{eval.synth}~(h), and hinder convergence.
%
The purely visual neural scene representation of NV~\cite{lombardiNeuralVolumesLearning2019} does not employ physical priors.
Despite encoding a representation over time by relying on inputs from the observed sequence for the inference of new views, it does not contain information about the motion between consecutive frames.
To compute volumetric samples that are comparable,
we extract a density by discarding black occlusions.
While the measurements for $\dens$ are in a similar range as our approach, the image space errors for $\imageRand$ are larger, despite not providing velocities.

Compared to all three previous methods, our global-transport approach outperforms the others in terms of image space reconstruction and  most volumetric metrics in \myreftab{eval}.
This is also highlighted in \myreffig{eval.realTar.5view}, where we compare a target frame qualitatively to a nearby target view.
Most importantly, our method reconstructs a temporally consistent motion over the full duration of the observed sequence -- a key property for retrieving a realistic fluid motion.
We highlight this behavior with an additional test in \myreffig{warp_test}, where an initial state is advected over the course of 120 steps with the optimized velocity sequence. Our approach yields the closest match with the reference configuration.

\begin{figure}
    \centering
        \mygraphicsB{width=0.195\linewidth}{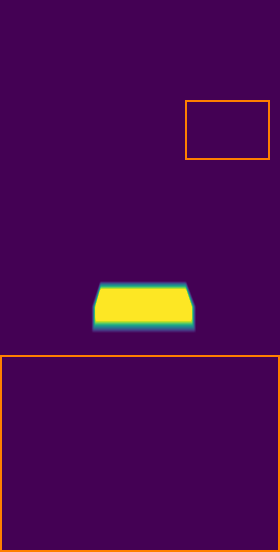}{a) Initial}%
        \hfill%
        \mygraphicsB{width=0.195\linewidth}{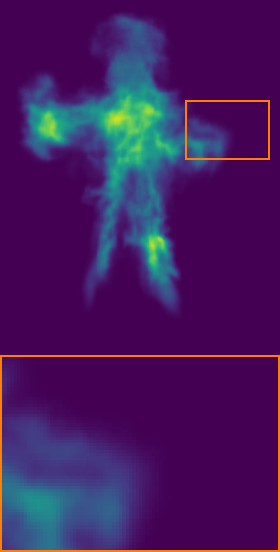}{b) TF \mycite{zangTomoFluidReconstructingDynamic2020}}%
        \hfill%
        \mygraphicsB{width=0.195\linewidth}{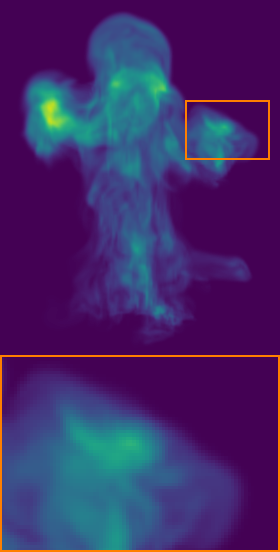}{c) SF \mycite{eckertScalarFlowLargescaleVolumetric2019}}%
        \hfill%
        \mygraphicsB{width=0.195\linewidth}{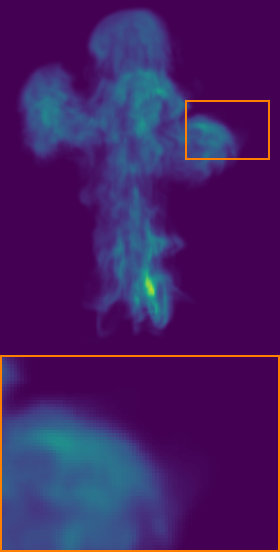}{d) \varGlobWarp}%
        \hfill%
        \mygraphicsB{width=0.195\linewidth}{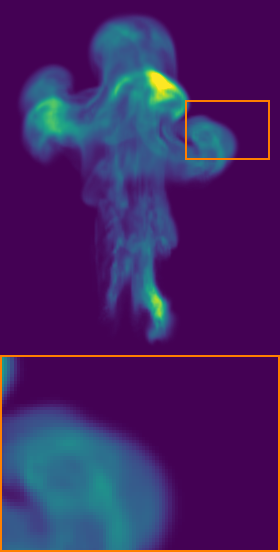}{e) Reference}%
    \caption{
    Evaluation of transport accuracy for an initial state (a) advected with the velocity reconstructions of TF (b) and SF (c). 
    Our result (d) is closest to the reference in (e). 
    }
    \label{fig:warp_test}
    \vspace{-0.3cm}
\end{figure}

\begin{figure}
        \centering%
        \mygraphicsTtiny{width=0.166\linewidth}{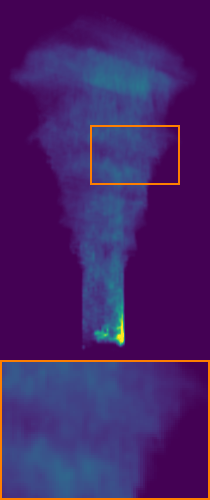}{a) \varForward}%
        \hfill%
        \mygraphicsTtiny{width=0.166\linewidth}{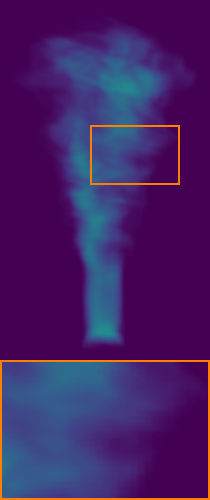}{b) \varCoupledMS}%
        \hfill%
        \mygraphicsTtiny{width=0.166\linewidth}{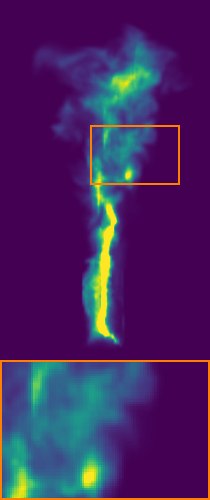}{c) \shortGobWarp}%
        \hfill%
        \mygraphicsTtiny{width=0.166\linewidth}{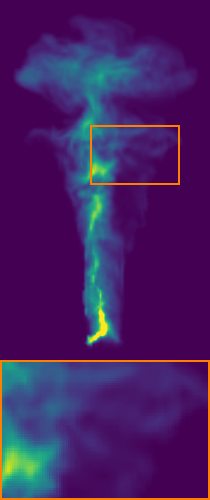}{d) \varDisc}%
        \hfill%
        \mygraphicsTtiny{width=0.166\linewidth}{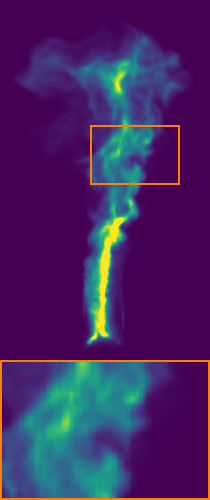}{e) \varDiscTemp~\mycite{xieTempoGANTemporallyCoherent2018}}%
        \hfill%
        \mygraphicsTtiny{width=0.166\linewidth}{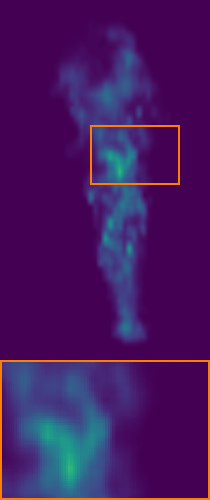}{f) pGAN~\mycite{henzlerEscapingPlatoCave2019}}%
        \newline%
        \includegraphics[width=0.166\linewidth]{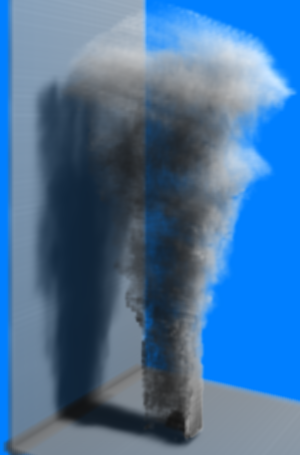}%
        \hfill%
        \includegraphics[width=0.166\linewidth]{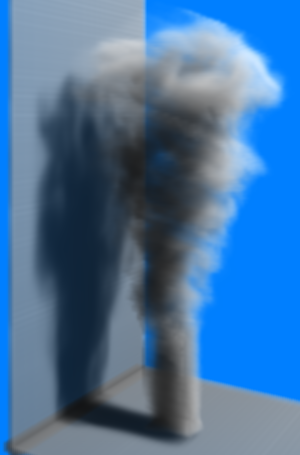}%
        \hfill%
        \includegraphics[width=0.166\linewidth]{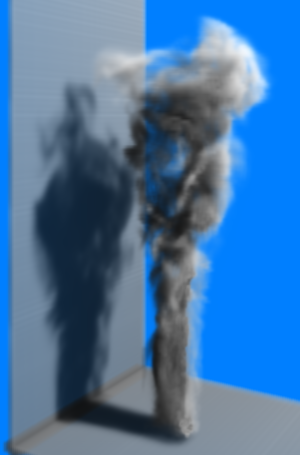}%
        \hfill%
        \includegraphics[width=0.166\linewidth]{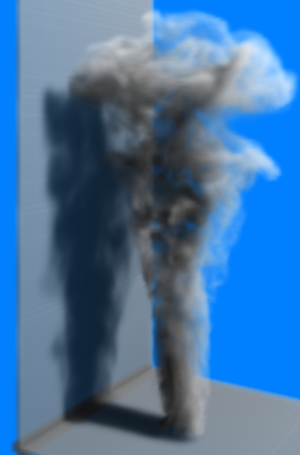}%
        \hfill%
        \includegraphics[width=0.166\linewidth]{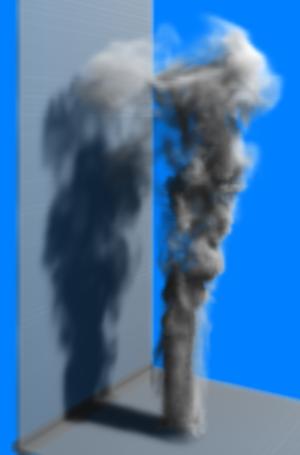}%
        \hfill%
        \includegraphics[width=0.166\linewidth]{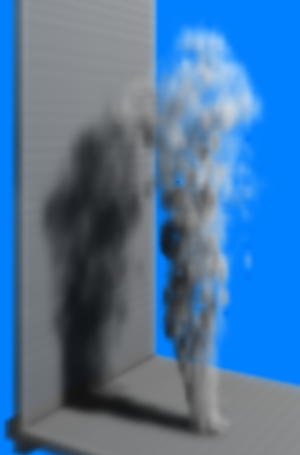}%
    \caption{
    Single-view reconstructions of a captured fluid (target at \degt{0}).
    Top \& middle: density from a \degt{90} side view.
    Bottom: rendered visualization at \degt{60}.
    Our version with discriminator (d) yields coherent and sharp flow structures, preventing any smearing along the unconstrained direction (a, b) or streak artefacts (c).
    Temporal supervision \mycite{xieTempoGANTemporallyCoherent2018} has no added benefit (e), while the purely learning-based method \mycite{henzlerEscapingPlatoCave2019} fails to generate coherent features (f).
    }
    \label{fig:eval.real.1view}
    \vspace{-0.2cm}
\end{figure}

\subsection{Single-View Reconstruction}\label{sec:eval.1view} 

We demonstrate the learned self-supervision, \ie, the efficacy of the discriminator loss, on a reconstruction of 120 frames from a real capture \mycite{eckertScalarFlowLargescaleVolumetric2019}. To evaluate the results from unconstrained directions we also include FID measurements~\mycite{heusel2017gans} in \myreftab{eval}~(b).

\myParSpace{}
\myParSpace{}
\paragraph{Ablation}

Results of the single-view ablation for an unseen view at a $90^{\circ}$ angle are shown in \myreffig{eval.real.1view}.
Here, the \textit{forward} variant (a) highlights the under-constrained nature of the single-view reconstruction problem with strong smearing along depth. 
The \textit{coupling} version (b) refines the reconstructed volumes, but still contains noticeable striping artifacts as well as a lack of details.
Our \textit{global transport} formulation (c) resolves these artifacts and yields a plausible motion, which, however, does not adhere to the motions observed in the input views. 
The \textit{full} version (d) introduces the methodology explained in \myrefsec{method.disc}, and arrives at our full algorithm including global transport and the discriminator network.
It not only sharpens the appearance of the plume, but also makes the flow more coherent by transferring information about the real-world behavior to unseen views. This is depicted in \myreffig{eval.realTar.1view} (c), where flow details closely match the style of an example view (d).

\begin{figure}%
    \centering%
    \mygraphicsB{height=0.34\linewidth}{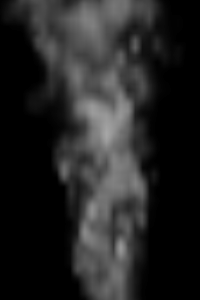}%
    {\shortstack[l]{a) pGAN~\mycite{henzlerEscapingPlatoCave2019}}}%
    \hfill%
    \mygraphicsB{height=0.34\linewidth}{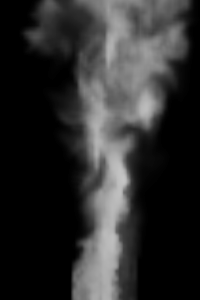}%
    {\shortstack[l]{b) \varDisc{} w/o disc. 
    }}%
    \hfill%
    \mygraphicsB{height=0.34\linewidth}{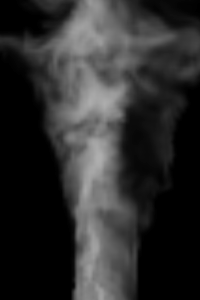}%
    {\shortstack[l]{c) \varDisc
    }}%
    \hfill%
    \mygraphicsB{height=0.34\linewidth}{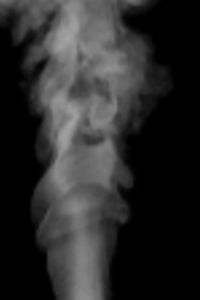}{d) Ex. view}%
    \caption{
    Unseen view of a single-view reconstruction: (a) previous work, our algorithm without (b) and with discriminator (c).
    The discriminator in (c) yields an improved reconstruction of small to medium scale structures in the flow that match the style of an example flow (d).
    }
    \label{fig:eval.realTar.1view}%
\end{figure}%

\myParSpace{}
\myParSpace{}
\paragraph{State of the Art}
Only few works exist that target volumetric single-view reconstructions over time. As two representatives, we evaluate the \textit{temporal discriminator} $D_t$ proposed in previous work for fluid super-resolution~\mycite{xieTempoGANTemporallyCoherent2018}, and the \textit{Platonic GAN} approach for volumetric single view reconstructions~\mycite{henzlerEscapingPlatoCave2019}.

The temporal self-supervision (Full-$D_t$)~\mycite{xieTempoGANTemporallyCoherent2018}
was proposed for settings without physical priors. We have extended our full algorithm with this method, and note that it does not yield substantial improvements for the optimized motion, \ef{see} \myreffig{eval.real.1view}~(e). This highlights the capabilities of the global transport, which already strongly constrains the solution. Thus, a further constraint in the form of a temporal self-supervision has little positive influence, and slightly deteriorates the FID results in \myreftab{eval}~(b).

The \textit{Platonic GAN} (pGAN) was proposed for learning 2D supervision per class of reconstructed objects. 
We have used this approach to pre-train a network for per-frame supervision of the rising smoke case of our single-view evaluation yielding improvements in terms of depth artifacts. However, due to the lack of constraints over time, the solutions are incoherent, and do not align with the input sequence, 
leading to an FID of more than 180.

In contrast, the solution of our full algorithm yields a realistically moving sequence that very closely matches the input view while yielding plausible motions even for highly unconstrained viewing directions. This is reflected in the FID measurements in \myreftab{eval}~(b), where our full algorithm outperforms the other versions with a score of 140.3.
We found the discriminator to be especially important for noisy, real-world data, where hard, model-based constraints can otherwise lead to undesired minima in the optimization process.
Both comparisons highlight their importance of learned self-supervision for single-view reconstructions, and their capabilities in combination with strong priors for the physical motion.

\subsection{Real Data Single-View Reconstructions}\label{sec:res.add}

We show two additional real flow scenarios reconstructed from single viewpoints with our full algorithm in \myreffig{eval.extra}.
For each case, our method very closely recovers the single input view, as shown in the top row of each case. In addition, the renderings from a new viewpoint in the bottom row highlight the realistic motion as seen from unconstrained viewpoints. 
Here, our optimization successfully computes over 690 million degrees of freedom\ef{, \ie, grid cells,} in a coupled fashion for the wispy smoke case.
Resolution details are given in \myrefapp{eval}, while the supplemental video shows these reconstructions in motion.
%
Interestingly, we found it beneficial to re-train the discriminator alongside each new reconstruction. Reusing a pretrained network yields inferior results, which indicates that the discriminator changes over the course of the reconstruction to provide feedback for different aspects of the solution.

\begin{figure}[t]%
    \centering%
    \begin{Overpic}{%
    \includegraphics[width=0.099\linewidth]{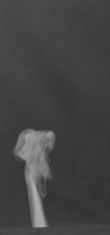}%
    \includegraphics[width=0.099\linewidth]{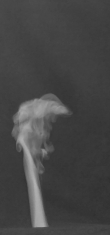}%
    \includegraphics[width=0.099\linewidth]{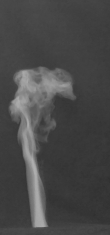}%
    \includegraphics[width=0.099\linewidth]{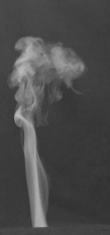}%
    \includegraphics[width=0.099\linewidth]{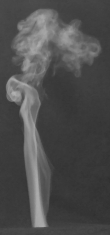}%
    }%
        \put(2,36){\begin{scriptsize}{\color{picCol}a) Target}\end{scriptsize}}%
    \end{Overpic}%
    \hfill%
    \begin{Overpic}{%
    \includegraphics[width=0.099\linewidth]{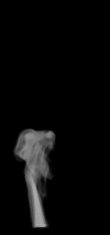}%
    \includegraphics[width=0.099\linewidth]{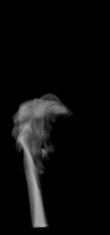}%
    \includegraphics[width=0.099\linewidth]{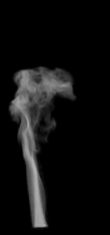}%
    \includegraphics[width=0.099\linewidth]{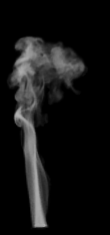}%
    \includegraphics[width=0.099\linewidth]{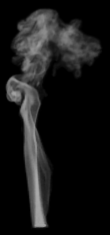}%
    }%
        \put(2,36){\begin{scriptsize}{\color{picCol}a) Reconstruction}\end{scriptsize}}%
    \end{Overpic}%
    \vspace{1.5pt}\newline%
    \begin{Overpic}{%
    \includegraphics[width=0.199\linewidth]{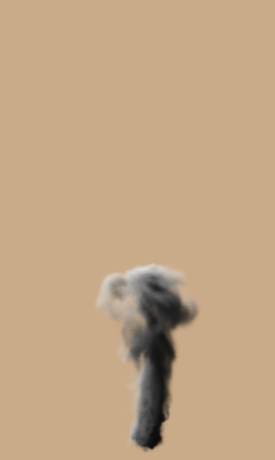}%
    \includegraphics[width=0.199\linewidth]{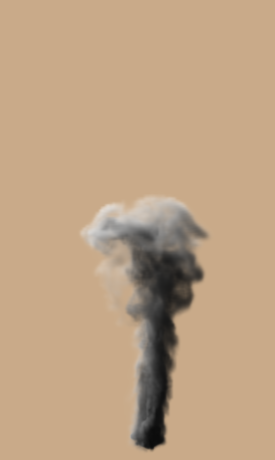}%
    \includegraphics[width=0.199\linewidth]{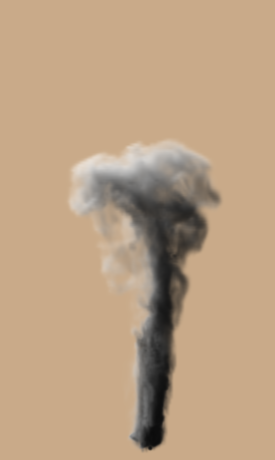}%
    \includegraphics[width=0.199\linewidth]{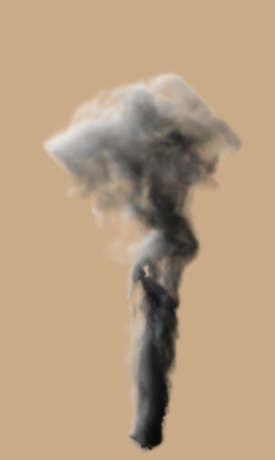}%
    \includegraphics[width=0.199\linewidth]{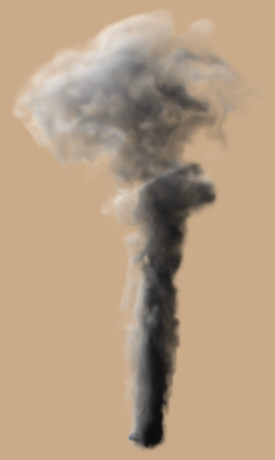}%
    }%
        \put(1,29.5){\begin{scriptsize}{a) Novel View}\end{scriptsize}}%
    \end{Overpic}%
    \vspace{1.5pt}\newline%
    \begin{Overpic}{%
    \includegraphics[width=0.099\linewidth]{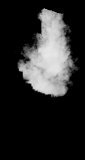}%
    \includegraphics[width=0.099\linewidth]{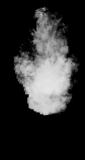}%
    \includegraphics[width=0.099\linewidth]{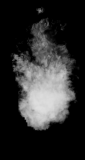}%
    \includegraphics[width=0.099\linewidth]{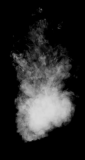}%
    \includegraphics[width=0.099\linewidth]{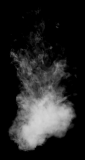}%
    }%
        \put(2,2.5){\begin{scriptsize}{\color{picCol}b) Target}\end{scriptsize}}%
    \end{Overpic}%
    \hfill%
    \begin{Overpic}{%
    \includegraphics[width=0.099\linewidth]{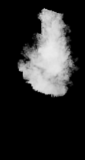}%
    \includegraphics[width=0.099\linewidth]{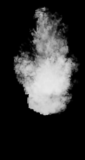}%
    \includegraphics[width=0.099\linewidth]{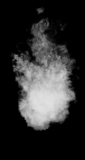}%
    \includegraphics[width=0.099\linewidth]{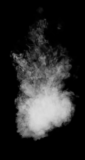}%
    \includegraphics[width=0.099\linewidth]{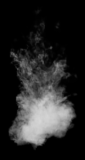}%
    }%
        \put(2,2.5){\begin{scriptsize}{\color{picCol}b) Reconstruction}\end{scriptsize}}%
    \end{Overpic}%
    \vspace{1.5pt}\newline%
    \begin{Overpic}{%
    \includegraphics[width=0.199\linewidth]{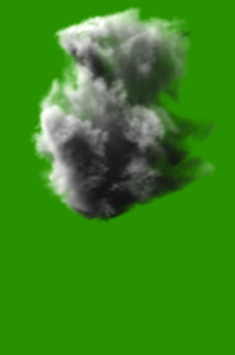}%
    \includegraphics[width=0.199\linewidth]{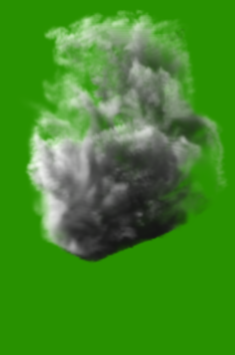}%
    \includegraphics[width=0.199\linewidth]{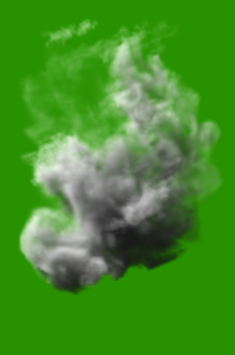}%
    \includegraphics[width=0.199\linewidth]{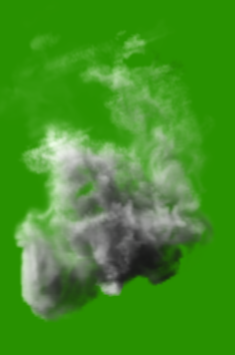}%
    \includegraphics[width=0.199\linewidth]{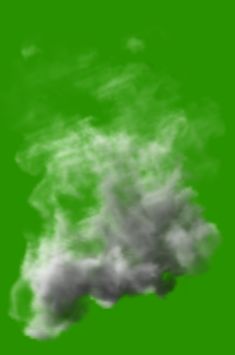}%
    }%
        \put(1,2){\begin{scriptsize}{b) Novel View}\end{scriptsize}}%
    \end{Overpic}%
    \caption{
    Additional single-view reconstructions. Top left: input views, top right: reconstruction from input view, bottom: different viewpoint with modified lighting. 
    Our algorithm successfully reconstructs realistic motions for wispy smoke (a), as well as thicker volumes (b).
    }%
    \label{fig:eval.extra}%
    \vspace{-0.2cm}
\end{figure}%

\section{Discussion and Conclusions}\label{sec:discuss}

We have presented a first method to recover an end-to-end motion of a fluid sequence from a single input view. This is achieved with a physical prior in the form of the global transport formulation, a differentiable volumetric renderer and a learned self-supervision for unseen angles.

Our \ef{single-view} method comes with the limitation that
the learned self-supervision assumes that the flow looks similar to the input view from unseen angles.
\ef{The discriminator can only make the results look plausible by transferring the visual style of the given references.
Physical correctness is induced by the direct, physical transport constraints.}
However, fluid flow naturally exhibits self similarity, and the discriminator performs significantly better than re-using the same input for constraints from multiple angles.
\ef{Dissipation effects are not modeled in our approach, and would thus lead to errors in $\Ldiv$ or $\Ldenswarp$.
We further do not support objects/obstacles immersed in the flow.}
Additionally, we rely on positional information about the inflow into the reconstruction domain. Jointly optimizing for inflow position and flow poses an interesting challenge for future work.
For our full method we have measured an average reconstruction time of $\sim30$ minutes per frame using a single 
\ef{GPU, more details are in \myrefapp{runtime}.}

In conclusion, our method makes an important step forward in terms of physics- and learning-based reconstruction of natural phenomena. We are especially excited about future connections to generative models and other physical phenomena, 
such as turbulent flows \cite{um2020sol}, liquids \cite{schenck2018spnets}, or even elastic objects\cite{weiss2020ssc}.


\section{Acknowledgements}\label{sec:acks}

This work was supported by the Siemens/IAS Fellowship \textit{Digital Twin}, and the ERC Consolidator Grant \textit{SpaTe} (CoG-2019-863850).


\clearpage
\appendix
\section*{Supplemental Document}
\ef{This is the supplemental material for the CVPR 2021 paper \emph{Global Transport for Fluid Reconstruction with Learned Self-Supervision}. The accompanying video as well as the source code can be found on the project website.}
\section{Details on Method and Implementation}

\subsection{Algorithm} \label{app:sup.alg}
Here we detail the different algorithms used in the ablation.
\myrefAlg{recon-dens-step} shows how the density of a single frame is reconstructed. This variant corresponds to the \emph{single} version using the single pass of \myrefsec{recon}. The \emph{forward} reconstruction is summarized in \myrefalg{recon-fwd}; it uses the density reconstruction of \myrefalg{recon-dens-step}.
In \myrefalg{recon-main} we depict the \emph{coupling} (\varCoupled) over time, and \myrefalg{recon-Gwarp} shows our full method with global transport optimization \emph{global transport} (Glob-Trans).

\begin{algorithm}
\Fn{\FnOptDensStep{$\dens, \Ldens$}}{
    $\light \algAssign$ \FnLight{$\dens$, lights}\;
    $\hat{\image}_{\iviews} \algAssign \render(\dens, \light, \iviews)$\;
    $\nabla\dens \algAssign \partial\Ldens / \partial\dens$\;
    Update $\dens$ with $\nabla\dens$ using Adam\;
    $\dens \algAssign$ \FnMax{$\dens, 0$}; //clip negative density\\
}
\Fn{\FnOptVelStep{$\dens, \Lvel$}}{
    $\nabla\vel \algAssign \partial\Lvel / \partial\vel$\;
    Update $\vel$ with $\nabla\vel$ using Adam\;
}
\Fn{\FnOptDiscStep{$\dens$}}{
    $\image_{r} \sim R $\;
    $f \sim \Omega$; //sample random views\\
    $\hat{\image}_{f} \algAssign \render(\dens, \light, f)$\;
    $\hat{\image}_{h} \sim \text{history}$\;
    $\text{history} \algAssign \text{history} \circ \hat{\image}_{f}$\;
    $\hat{\image}_{f} \algAssign \hat{\image}_{f} \circ \hat{\image}_{h}$\;
    Update $\Theta_{\disc}$ with $\pderInl{\Ldisc(\dens,1)}{\Theta_{\disc}}$; //\myrefeq{loss.disc}\\
}
\caption{Update Steps}\label{alg:recon-dens-step}
\end{algorithm}

\begin{algorithm}
    \Fn{\FnOptDens{$\dens$}}{
        \For{n}{
            \FnOptDensStep{$\dens, \Ltarget$}\;
        }
    }
    \Fn{\FnOptVel{$\vel, \dens_{from}, \dens_{to}, n_{\text{MS}}$}}{
        \For{$i=0$ \KwTo $n$}{
            \If{$i \in n_{\text{MS}}$}{
                $\vel \algAssign$ \FnResizeGrid{$\vel$}\;
            }
            \FnOptVelStep{$\vel, \Ldwarp + \Ldiv$}\;
        }
    }
    \Fn{\FnOptFwd{$\{\dens\},\{\vel\}$}}{
        $\dfirst \algAssign \hull^0 \cdot c$ \;
        $\vfirst \algAssign \hull^1 \cdot$ rand\;
        \FnOptDens{$\dfirst$}; //\myrefalg{recon-dens-step}\\
        $\dens^1 \algAssign \dens^0$\;
        \FnOptDens{$\dens^1$}\;
        \FnOptVel{$\vel^0, \dens^0, \dens^1, n_{\text{MS}}$}\;
        \For{$t=1$ \KwTo $\nframes-1$}{
            $\vcurr \algAssign \warp(\vprev, \vprev)$\;
            $\dnext \algAssign \warp(\dcurr, \vcurr)$\;
            \FnOptDens{$\dnext$}\;
            \FnOptVel{$\vcurr, \dcurr, \dnext, \emptyset$}\;
        }
        $\vel^{\nframes-1} \algAssign \warp(\vel^{\nframes-2}, \vel^{\nframes-2})$\;
    }
    \caption{Forward Pass}\label{alg:recon-fwd}
\end{algorithm}

\begin{algorithm}
    \FnOptFwd{$\{\dens\},\{\vel\}$}\;
    \For{$i=0$ \KwTo $n$}{
        \For{$\iframes=0$ \KwTo $\nframes$}{
            \FnOptDensStep{$\dcurr, \Ltarget + \Ldwarp + \Ldisc$}\;
            \FnOptVelStep{$\vel, \Ldwarp + \Lvwarp + \Ldiv$}\;
        }
        \If{\VarDiscActive}{
            \FnOptDiscStep{$\{\dens\}$}\;
        }
    }
    \caption{Coupled Optimization}\label{alg:recon-main}
\end{algorithm}

\begin{algorithm}
    \FnOptFwd{$\{\dig\},\{\vel\}$}\;
    $\dfirst \algAssign \gstep{\dig}{0}$\;
    \For{$i=0$ \KwTo $n$}{
        \If{$i \in n_{\text{MS}}$}{
            $\dfirst \algAssign$ \FnResizeGrid{$\dfirst$}\;
            \For{$\iframes=\ffirst$ \KwTo $\nframes$}{
                $\vcurr \algAssign$ \FnResizeGrid{$\vcurr$}\;
            }
        }
        // build sequence from first frame $\dens^0$\\
        \For{$\iframes=1$ \KwTo $\nframes$}{
            $\dcurr \algAssign \warp(\dprev, \vprev)$\;
        }
        $\nabla \gstep{\dens}{\nframes}_{\warp} \algAssign 0$; // init gradient EMA\\
        \For{$\iframes=\flast$ \KwTo $-1$}{
            $\nabla \dcurr \algAssign \partial\Ldens(\dcurr)/\partial \dcurr$\;
            $\nabla \dcurr_{\warp} \algAssign  \partial\warp(\dcurr, \vcurr)/\partial\dcurr \cdot \nabla \dnext_{\warp}$\;
            \If{$\iframes==\ffirst$}{
                Update $\dcurr$ with $\nabla \dcurr + \lambda_{\dens_{\warp}} \nabla \dcurr_{\warp}$\;
            }
            $\nabla \vcurr_{\warp} \algAssign \partial\warp(\dcurr, \vcurr)/\partial\vcurr \cdot \nabla \dnext_{\warp}$\;
            Update $\vcurr$ with $\nabla \vcurr + \lambda_{\vel_{\warp}} \nabla \vcurr_{\warp}$\;
            $\nabla \dcurr_{\warp} \algAssign \beta * \nabla \dcurr_{\warp} + (1 - \beta) * \nabla \dcurr$\;
        }
        \If{\VarDiscActive}{
            \FnOptDiscStep{$\{\dens\}$}\;
        }
    }
    \caption{Full Method}\label{alg:recon-Gwarp}
\end{algorithm}

\subsection{Discriminator}\label{app:impl.disc}

Our discriminator $\disc$ has a fully convolutional architecture
with 14 convolution layers (filters, strides):
(8,1)
(16,2) (16,1)
(24,2) (24,1)
(32,2) (32,1) (32,1)
(64,2) (64,1) (64,1)
(16,1) (4,1) (1,1).
All layers use a filter size of
$4 \times 4$ and are followed by LReLU activations\ef{, with leak of 0.2}.
We use SAME padding, but instead of padding with zeros we use the mirrored input to avoid issues described in \mycite{islamHowMuchPosition2020}.
For data augmentation we apply random crop, scale, rotation, intensity and gamma scale.
We keep a history of rendered fake samples $\hat{\image}_{f}$ to reuse as additional fake samples later for training the discriminator which improves the quality of the reconstruction and reduces rendering operations.
The batch size for each discriminator or density update is 8 real samples and 4 rendered fake samples, with 4 additional fake samples from the history when training the discriminator.
The 'average' formulation of the relativistic GAN \mycite{jolicoeur-martineauRelativisticDiscriminatorKey2018a} allows to use different resolutions and batch sizes for real and fake samples, while
the least-squares variant gives empirically better results in our case.
We also use an $L_2$ regularization for the weights of the discriminator $\Theta_{\disc}$.

\ef{%

\subsection{Differentiable Warping} \label{app:impl.warp}
For the discretized operator $\warp$ we use a second-order transport scheme in order to preserve small-scale content on the warped fields \cite{selleUnconditionallyStableMacCormack2008}.
This MacCormack advection is based on Semi-Langrangian advection $\warp_{SL}$, where the advected quantity $\gstep{s}{\fnext}$ at position $\vec{x}$ is the tri-linear interpolation of $\gstep{s}{\iframes}$ at $\vec{x} - \vec{u} * \Delta t$. With this, the MacCormack-advection is defined as
\begin{equation}
	\begin{aligned}
		\hat{s}^{t+1} &:= \warp_{SL}(s^t, \vel^t),\\
		\hat{s}^{t} &:= \warp_{SL}(\hat{s}^{t+1}, - \vel^t),\\
		s^{t+1} &:= \hat{s}^{t+1} + 0.5 (s^t - \hat{s}^{t}).
	\end{aligned}
\end{equation}
However, we use a variant that reverts to $\warp_{SL}$ when the corrected term $s^{t+1}$ would exceed the values used for interpolation in the first $\warp_{SL}$:
\begin{equation}
	\begin{aligned}
		s^{t+1} :=
		\begin{cases}
			s^{t+1} & \text{if}~\hat{s}^{t+1}_{min} \le s^{t+1} \le \hat{s}^{t+1}_{max} \\
			\hat{s}^{t+1} &\text{else}
		\end{cases}.
	\end{aligned}
\end{equation}
This reversion is necessary as the un-clamped MacCormack correction causes issues like negative density and escalating inflow from the open boundaries.
The result is smoother than the one obtained without clamping, but still more detailed than $\warp_{SL}$.

The advected generic quantity $s$ can be both $\dens$ and $\vel$, in case of $\vel$ the components are handled individually and treated as scalars.

\myParSpace{}%
\myParSpace{}%
\paragraph{Differentiation}
We implemented a differentiable SL advection as tensorflow CUDA operation. Since the MacCormack advection is composed of two SL steps and a condition, its gradients are simply handled by automatic differentiation.
Since we optimize $\dens$ and $\vel$, both $\pderInl{\warp(s,\vel)}{s}$ and $\pderInl{\warp(s,\vel)}{\vel}$ are required.
For $\pderInl{\warp(s,\vel)}{s}$, the gradients
are just scattered to the grid positions used for the interpolation in the forward step, after being multiplied with the interpolation weights.
$\pderInl{\warp(s,\vel)}{\vel}$ is $\warp(\nabla s,\vel)$, \ie, during back-propagation the gradients are multiplied component wise with the advected spatial gradients of the scalar grid $s$.

\subsubsection{Boundary Handling}\label{app:bounds}
We use open boundaries for both density and velocity.
Technically that means that velocities at the boundaries are not constrained (or handled differently from the inner velocities in any way). In the advection, look-ups that fall outside of the volume are clamped to the boundaries, thus using the values there. Their gradients are also accumulated at the boundaries accordingly.

\subsubsection{Inflow}
Inflow is only handled explicitly for the density, the velocity has to solve all inflow via the open boundaries.
As we target rising smoke plumes we place an inflow region at the lower end of the visual hull $\hull$, with some overlap, although it could be composed of arbitrary regions of the volume without changing the algorithm (we use a simple mask to indicate inflow cells).
These cells are additional degrees of freedom and are optimized alongside the density. They are individual for every frame, and still used when using our global transport formulation.
The advection then becomes $\warp(\dcurr + \textnormal{inflow}, \vcurr) = \dnext$ and is used as such in all losses and the global transport. This is also how the inflow receives gradients.
The only constraint on the inflow is that $\dcurr + inflow \geq 0$, meaning that the inflow itself can be (partially) negative, and thus serve as outflow.
Density can also enter and leave the volume via the boundaries, just like the velocity.

\subsection{Differentiable Rendering} \label{app:impl.render}
For rendering we use a discrete ray-marching scheme that essentially replaces the integrals in \myrefeq{render} with discrete sums, stepping though the volume along the view-depth ($z$) with a fixed step size:
\begin{equation}
\begin{aligned}
    \sum_{z=0}^{Z} \light(x_z) e^{-\sum_{a=0}^{x_z} \dens(a)},
    e^{-\sum_{z=0}^Z \dens(x_z)}.
\end{aligned}
\end{equation}

To implement the single-scattered self-shadowing we use a shadow volume, instead of casting shadow rays from every sample point. Thus, the shadow computation is like rendering the transparency from the point of view of the light, but storing every $z$-slice to create the shadow volume.

We implement our custom ray-marching kernel and its analytic gradient as tensorflow CUDA operation. The backwards pass inside this operation is what standard auto-differentiation would yield: go backwards along the ray and scatter gradients to the locations used for interpolation in the forward pass.

\subsubsection{Sampling Gradients}\label{sec:abl.gradsample}
The naive back-propagation of gradients through the grid-sampling, \ie, scattering the gradients of a sample point to all used interpolants, weighted with the original interpolation weights, yields 
regular artifacts in the reconstruction,
likely due to the regular grid sampling. 
Using an intermediate grid allows us to invert the sampling for back-propagation with another sampling operation as in the forward pass, which can then employ mip-mapping, and thus reduces aliasing in both directions. Conceptually, this approach does not differentiate the sampling, but rather transforms the volume of gradients into another space, resulting in spatially smooth gradients. It is, however, very memory intensive as all samples have to be stored.
An equivalent gradient can also be obtained with scattering, by normalizing the gradients with their accumulated interpolation weights. This also avoids aliasing, as long as the rendering resolution is large enough, and is the method we use for our final results.
}

\subsubsection{\ef{Comparison to Linear Image Formation}}
The comparisons in \myreffigTwo{eval.render}{eval.render.supp} highlight the importance of the non-linear IF for visible light capturing. While a linear model reduces densities to account for effects like shadowing, the differentiable rendering in our optimization yields correct gradients to recover the original density, (b) \vs (c) in \myreffig{eval.render}.
\begin{figure}
    \centering
        \mygraphicsB{width=0.24\linewidth}{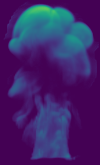}{a) Linear IF}%
        \mygraphicsB{width=0.24\linewidth}{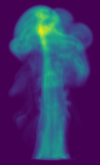}{b) Ours, Nl-IF}%
        \hfill
        \mygraphicsB{width=0.24\linewidth}{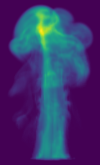}{c) Target $\dens$ vis.}%
        \mygraphicsB{width=0.24\linewidth}{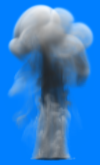}{d) Target view}%
    \caption{
    Non-linear IF optimization:
    a linear model \myquote{bakes} lighting into the density reconstruction (a), while our pipeline can recover the correct density distribution (b) via a differentiable renderer. (c,d) show target density as visualization and with lighting, respectively.
    }
    \label{fig:eval.render}
\end{figure}
\begin{figure}
    \centering
        \mygraphicsB{width=0.24\linewidth}{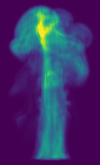}{a) recon BL}%
        \mygraphicsB{width=0.24\linewidth}{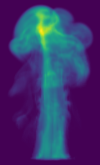}{b) recon LIN}%
        \hfill%
        \mygraphicsB{width=0.24\linewidth}{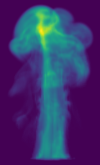}{c) target LIN}%
        \mygraphicsB{width=0.24\linewidth}{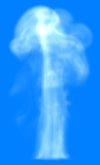}{d) target LIN}%
    \caption{A target rendered with a linear (LIN) IF can be recovered by the attenuated model (BL), as long as sufficient lighting is provided.
    }
    \label{fig:eval.render.supp}
\end{figure}

\ef{
\subsubsection{Comparison to Path-tracer} \label{app:sup.rendercmp}
We compare our renderer (\myrefsec{method.render}) to a path-traced reference in \myrefFig{render_cmp}.
Using 0 light bounces in the path-tracer (\ie only single-scattering shadow rays) produces results very similar to our renderer, while the results with more bounces (multi-scattering) change gracefully.
\begin{figure}%
     \centering%
     \includegraphics[width=0.28\linewidth]{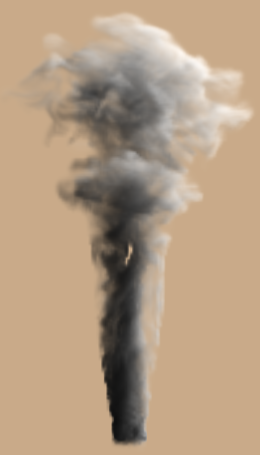}%
     \hfill%
     \includegraphics[width=0.28\linewidth]{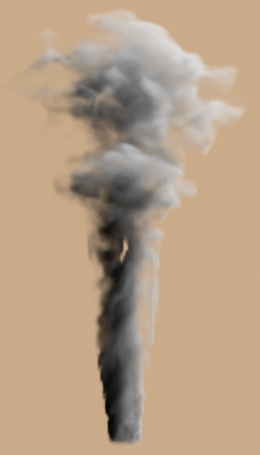}%
     \hfill%
     \includegraphics[width=0.28\linewidth]{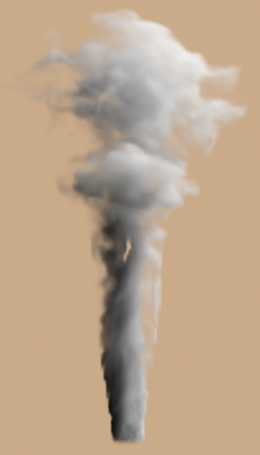}%
     \caption{\ef{Our renderer vs. Blender 2.91 \mycite{blender} Cycles path-tracer (128 SPP, denoised), 0 and 4 light bounces.}}%
     \label{fig:render_cmp}%
\end{figure}


}

\subsubsection{Backgrounds and Lighting Optimization}\label{app:sup.bkg-lightopt}
When using a black background or monochrome target image there is an ambiguity between high light intensity in conjunction with low density and lower light intensity with more density
as these two cases cannot be distinguished in a target image.
With a black background the transparency does not matter, and with a monochrome target (with any background) contributions from reflected light and background are indistinguishable.
Using a background with a color distinguishable from the smoke, the transparency, and therefore density, along a ray can be determined successfully.

The reconstruction consequently yields better results when using a colored background, as seen in \myreffig{render.bkg}.
However, as our real target dataset \mycite{eckertScalarFlowLargescaleVolumetric2019} uses only grey-scales, we likewise used black backgrounds for our synthetic tests.

We also experimented with optimizing the light intensities ($i_a$ and $i_p$),
which worked better with colored backgrounds. 
Because this adds additional degrees of freedom, making the reconstruction harder, we
ultimately determined our light intensities empirically.
\begin{figure}%
    \centering%
    \mygraphicsB{width=0.32\linewidth}{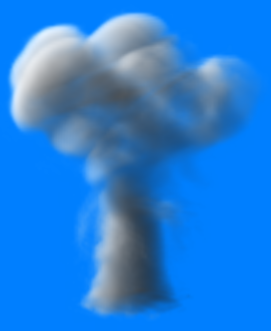}{a) Black Background}%
    \hfill%
    \mygraphicsB{width=0.32\linewidth}{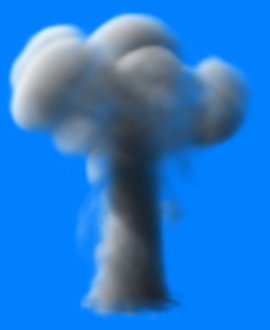}{b) Blue Background}%
    \hfill%
    \mygraphicsB{width=0.32\linewidth}{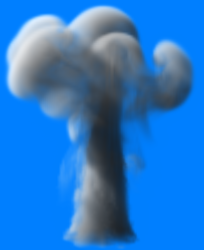}{c) Reference}%
    \caption{Side view (\degt{90}) of a density reconstructed with black (a) and light blue RGB(0,127,255) (b) background. A colored background helps to recover the correct density in shadowed regions.}%
    \label{fig:render.bkg}%
\end{figure}%

\ef{

\subsection{Visual Hull} \label{app:sup.hull}
For our multi-view reconstruction experiments with 5 target views, the visual hull $\hull$ is constructed only from the 5 available views, as described in \myrefsec{method.other}. After the background subtraction the target images are turned into a binary mask, using  a threshold of $\epsilon = 0.04$ to cut of residual noise. These masks are slightly blurred with $\sigma = 1$ (in pixels) before being
projected into the volume \myrefeqshort{hull}. The hulls are then blurred again with $\sigma = 0.5$ (in cells).

To construct a visual hull from a single view, we create 4 additional, evenly spread views from the available view by rotating it around the central y-axis (up) of the volume. The masks created from these auxiliary views are additionally mirrored at the projected rotation axis to reduce cut-offs in the original view when constructing the volume hull by intersection.
This works well for relatively symmetric rising plumes, but is not guaranteed to work for arbitrary shapes.
Using a visual hull to guide the densities noticeably improves the quality of the reconstruction, but constructing a hull for sparse or single-view reconstructions that adheres to physically correct motion remains a topic for future work.

\subsection{Multi-Scale Approach}\label{app:ms}
We employ a multi-scale approach during reconstruction, increasing the grid resolution while the spatial size of the domain stays the same.
The up-sampling is done using tri-linear interpolation.
In our full method we use a scaling factor per step of 1.2. Thus, for a reference resolution of 128 (see also \myrefapp{eval}), the velocity of the first frame is scaled 4 times
($14 \rightarrow 18 \rightarrow 21 \rightarrow 25 \rightarrow 30$) in the pre-optimization,
and density and velocity synchronously in the coupled phase an additional 8 times
($30 \rightarrow 36 \rightarrow 43 \rightarrow 51 \rightarrow 62 \rightarrow 74 \rightarrow 88 \rightarrow 106 \rightarrow 128$).
As the first frame velocity optimization operates on a lower resolution than the densities $\dcurr$ and $\dnext$ used therein, the densities are temporarily down-sampled (with filtering) to match the resolution of the velocity.

\subsection{Gradient Descent with Global Transport}
Naive automatic differentiation of a full 120 frame sequence via back-propagation would require a lot of memory as all intermediate results of the forward pass have to be recorded for the backwards pass.
To reduce the memory requirements we split the sequence to only do automatic back-propagation for one frame at a time. The back-propagation between frames is handled more explicitly (see also line 19 in \myrefAlg{recon-Gwarp}). The result is the same, at least when using \myrefeq{method.transport.backwarp} instead of the EMA version \myrefeqshort{method.transport.ema}.
We keep all resources in (CPU) RAM, only the current frame and intermediate steps needed for automatic differentiation are moved to the GPU.
} 

\subsection{Hyperparameters}\label{app:loss}
The hyperparameters, which can be set individually for the first frame in the pre-pass, the remaining frames of the pre-pass, and the main optimization, are reported in \myreftab{loss-bal}.
Some benefit from a linear ($\xrightarrow{lin}$) or exponential ($\xrightarrow{exp}$) fade-in or growth. We do not use any smoothness regularization for density or velocity.
\begin{table}[]
    \centering
    \begin{footnotesize}%
    \begin{tabular}{|c|c|c|c|}
        \hline
        Loss \textbackslash Pass & Fwd first & Fwd & Main \\\hline
        \hline%
        \multicolumn{4}{|c|}{Density} \\\hline%
        $n$ & 600 & 600 & 4200 \\\hline%
        $n_{\text{MS}}$ & - & - & 400 $\times$ 8\\\hline%
        $\eta_{\dens}$ & 3 & 3 $\xrightarrow{lin}$ 1 & 2.4 \\\hline%
        $\Ltarget$ & \multicolumn{3}{c|}{1.74e-5} \\\hline
        $\Ldwarp$ & 0 & 0 & 2.7e-10 $\xrightarrow{lin}$ 5.4e-10\\\hline
        $\Ldisc(\dens, -1)$ & 0 & 0 & 1.5e-5\\\hline
        \hline%
        \multicolumn{4}{|c|}{Velocity} \\\hline%
        $n$ & 6000 & 600 & 4200 \\\hline%
        $n_{\text{MS}}$ & 1000 $\times$ 4 & - & 400 $\times$ 8\\\hline%
        $\eta_{\vel}$ & 0.04 & 0.02 & $\xrightarrow{exp}$ 0.016\\\hline%
        $\Ldwarp$ & \multicolumn{3}{c|}{4.1e-10} \\\hline
        $\Lvwarp$ & 0 & 0 & 4e-11 $\xrightarrow{lin}$ 8e-11\\\hline
        $\Ldiv$ & 8.6e-10 $\xrightarrow{exp}$ & 2.6e-9 & $\xrightarrow{exp}$ 1.7e-8\\\hline
        \hline%
        \multicolumn{4}{|c|}{Discriminator$^\dagger$} \\\hline%
        $\eta_{\vel}$ & - & - & 2e-4\\\hline%
        $|\Theta_{\disc}|^2$ & - & - & 2e-3\\\hline%
    \end{tabular}
    \end{footnotesize}%
    \caption{Hyperparameters for the first (lines 12 and 15 in \myrefalg{recon-fwd}) and remaining frames of the forward reconstruction and our coupled global transport optimization (\myrefalg{recon-Gwarp}). learning rate $\eta$, iterations $n$ with multi-scale intervals $n_{\text{MS}}$ with a resize factor of 1.2.
    }\label{tab:loss-bal}%
\end{table}
%


\ef{
\subsection{Hardware and Runtime Statistics} \label{app:runtime}
We used the following hard- and software configuration for our experiments:
Hardware:
CPU: Intel(R) Core(TM) i7-6850K,
GPU: Nvidia GeForce GTX 1080 Ti 11GB,
RAM: 128GB;
Software (versions):
Python 3.6.9,
Tensorflow 1.12 (GPU),
CUDA 9.2.
We implemented custom tensorflow operation in CUDA for rendering (ray-marching and shading) and advection as well as their gradient operations.

For our full method, using the resolution detailed in \myrefapp{eval}, we have measured an average reconstruction time 
of 43 - 65 hours ($\sim30$ minutes per frame, variation based on system load) for 120 frames of single view reconstruction from real targets with our current implementation and hardware.
The pre-optimization \myrefsecshort{recon} takes only 2\% of the total time.
The majority of the time is spent on the density optimization (60\%, 38\% of which is needed for the discriminator loss and rendering the needed fake images), while 
Calculating and back-propagating though the losses for $\vel$ makes up 30\% of the time.
The manual back-propagation though the global transport via \myrefeq{method.transport.ema} takes 4\% of the total runtime.
} 

\begin{figure*}[t!]
    \centering
        \mygraphicsT{width=0.11\linewidth}{images/dens-vis_f120-1_frame125/00_single_dens_side.png}{a) \varSingle}%
        \mygraphicsT{width=0.11\linewidth}{images/dens-vis_f120-1_frame125/02_fwd-warp_dens_side.png}{b) \varForward}%
        \mygraphicsT{width=0.11\linewidth}{images/dens-vis_f120-1_frame125/03_coupled_dens_side.png}{c) \varCoupled}%
        \mygraphicsT{width=0.11\linewidth}{images/dens-vis_f120-1_frame125/04_coupled-grow_dens_side.png}{d) \varCoupledMS}%
        \mygraphicsT{width=0.11\linewidth}{images/dens-vis_f120-1_frame125/05_full_dens_side.png}{e) \varGlobWarp}%
        \mygraphicsT{width=0.11\linewidth}{images/dens-vis_f120-1_frame125/99_synthGT_dens_side.png}{f) Reference}%
        \hfill%
        \mygraphicsT{width=0.11\linewidth}{images/dens-vis_f120-1_frame125/20_scalarFlow_dens_side.png}{g) SF \mycite{eckertScalarFlowLargescaleVolumetric2019}}%
        \mygraphicsT{width=0.11\linewidth}{images/dens-vis_f120-1_frame125/22_tomoFluid_dens_side.png}{h) TF \mycite{zangTomoFluidReconstructingDynamic2020}}%
        \mygraphicsT{width=0.11\linewidth}{images/dens-vis_f120-1_frame125/21_neuralVoluems_dens_side.png}{i) NV \mycite{lombardiNeuralVolumesLearning2019}}%
        \newline
        \includegraphics[width=0.11\linewidth]{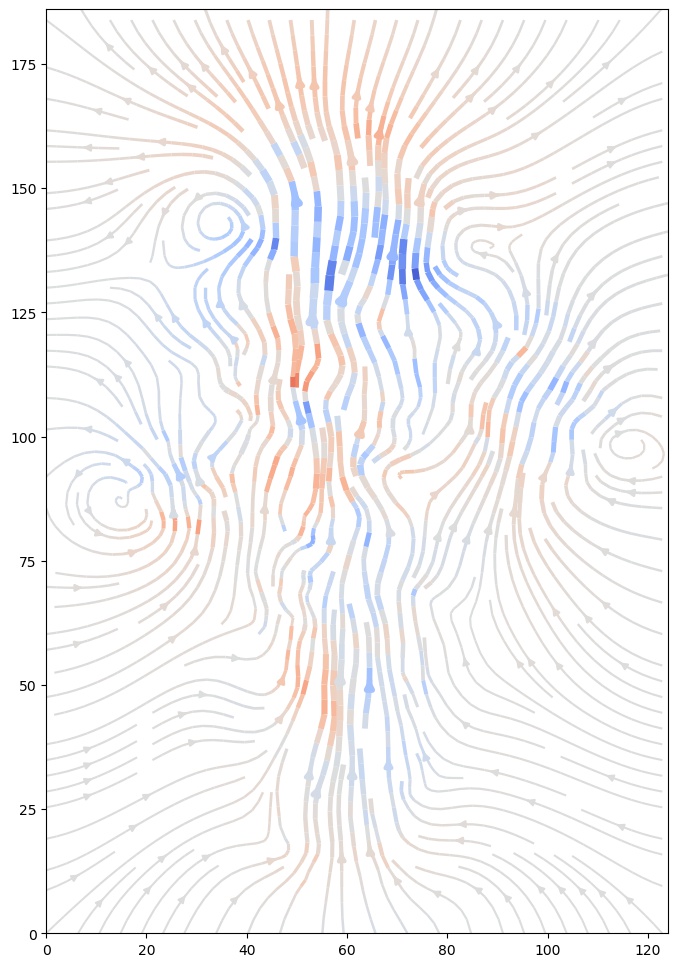}%
        \includegraphics[width=0.11\linewidth]{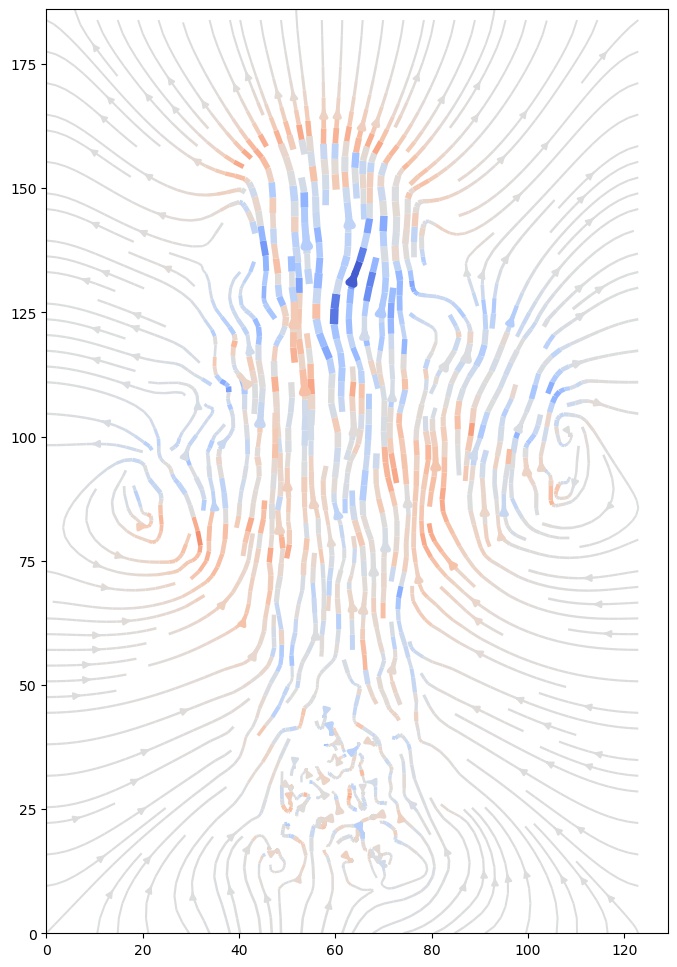}%
        \includegraphics[width=0.11\linewidth]{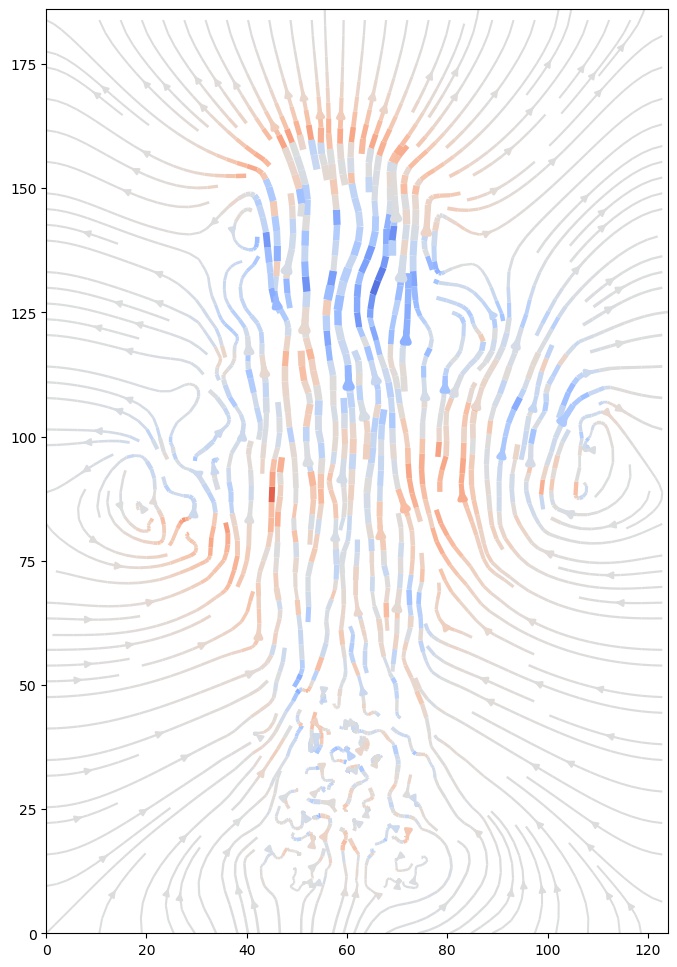}%
        \includegraphics[width=0.11\linewidth]{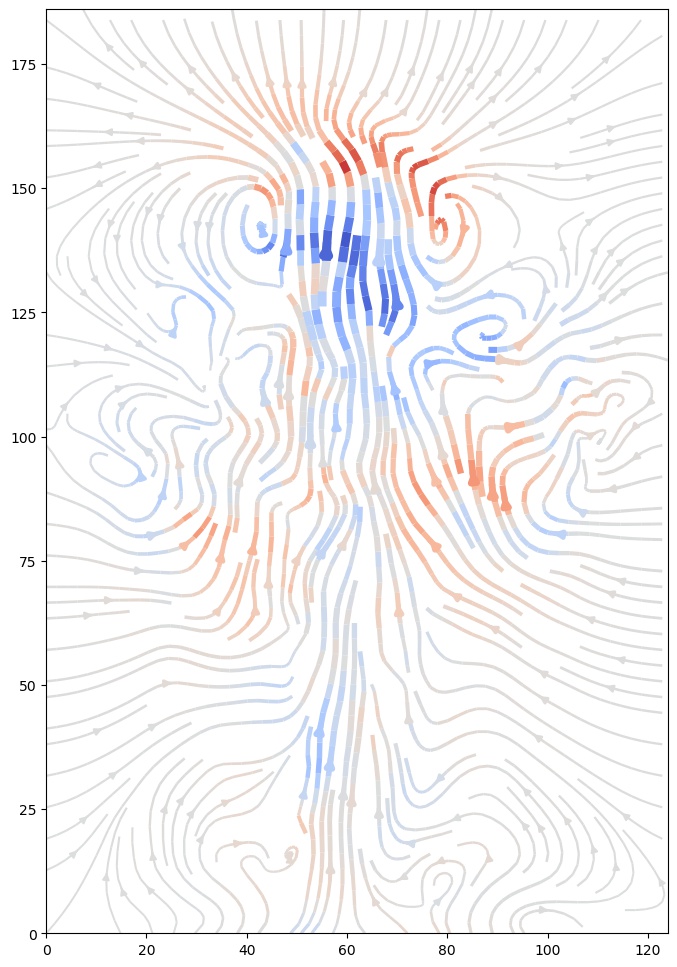}%
        \includegraphics[width=0.11\linewidth]{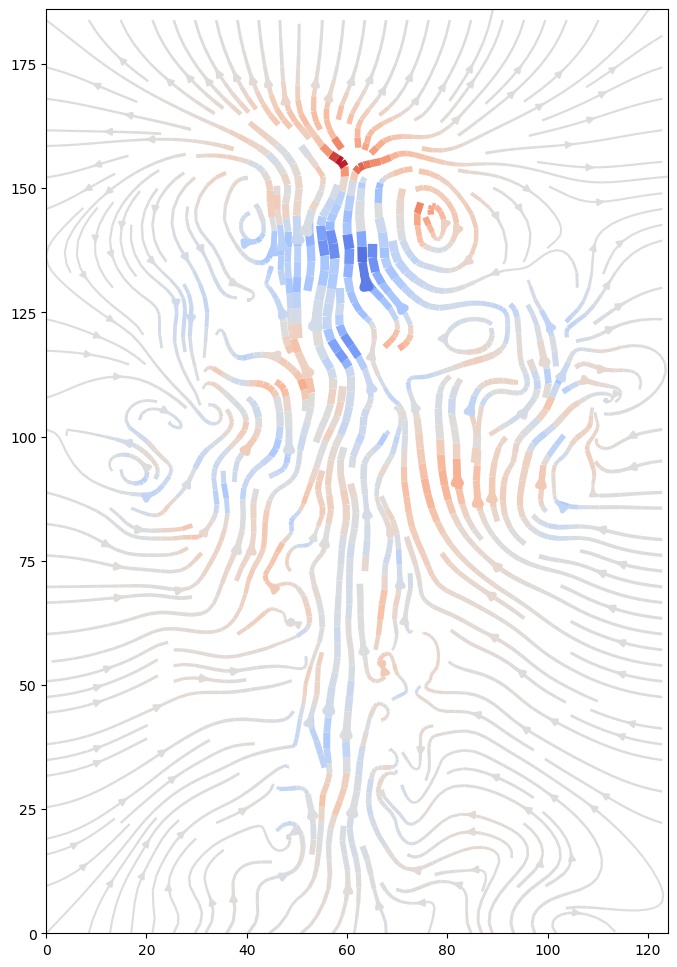}%
        \includegraphics[width=0.11\linewidth]{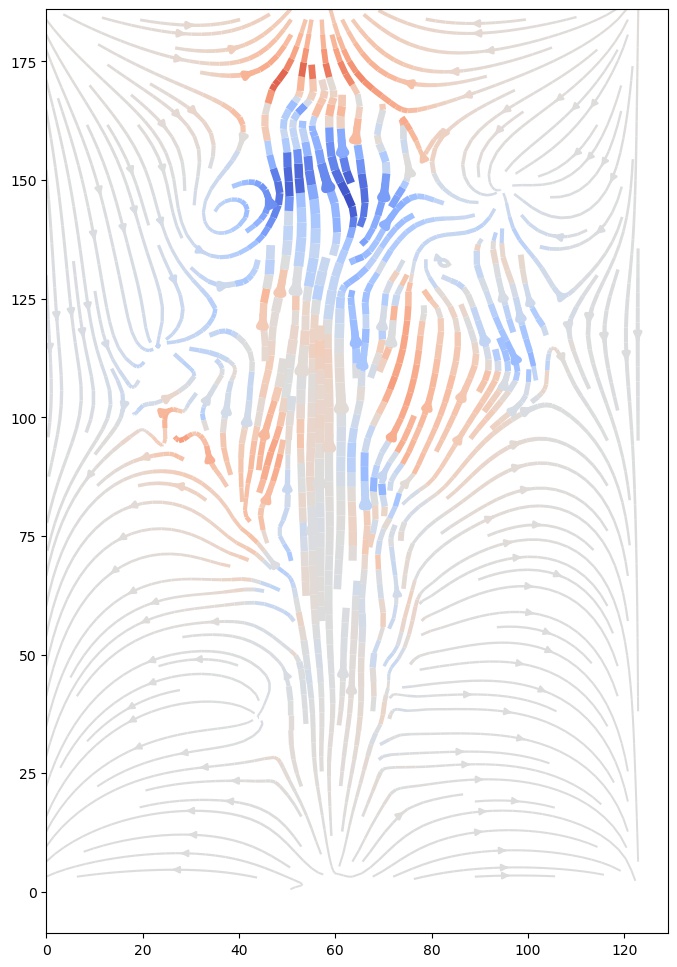}%
        \hfill%
        \includegraphics[width=0.11\linewidth]{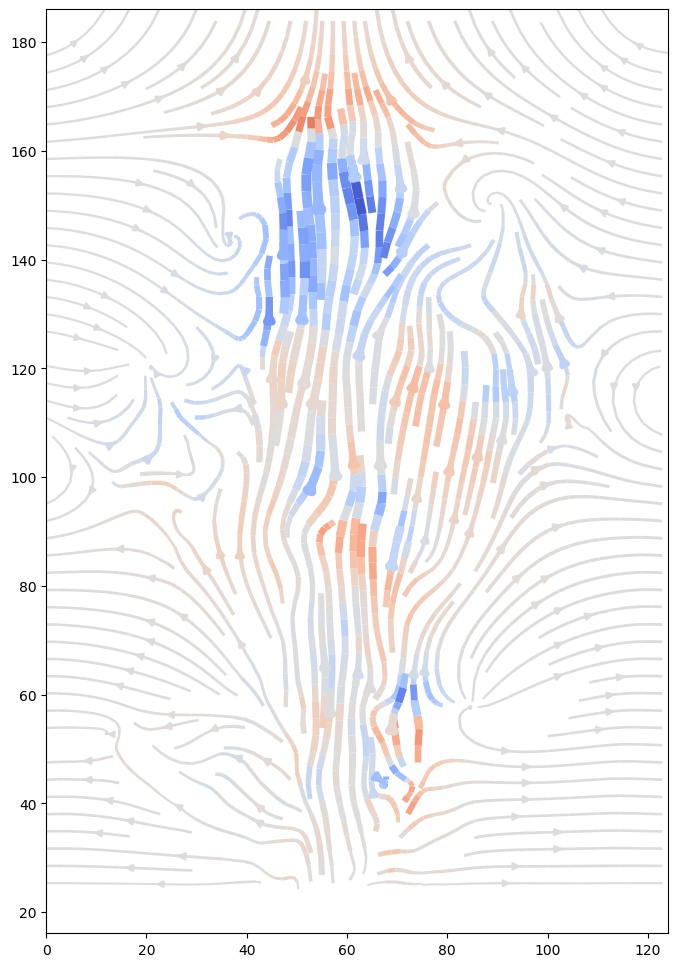}%
        \includegraphics[width=0.11\linewidth]{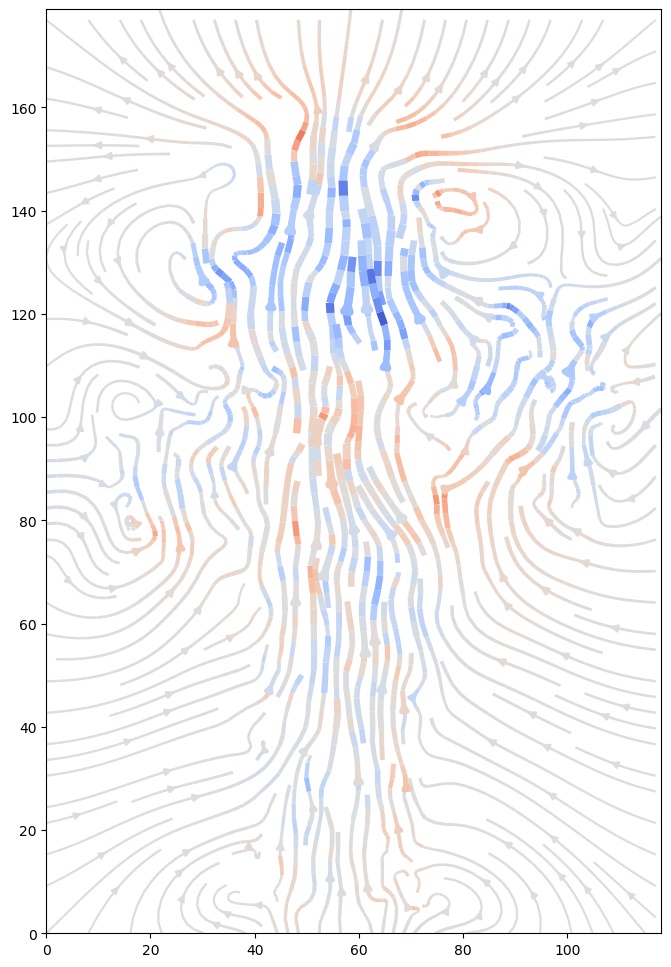}%
        \includegraphics[width=0.11\linewidth]{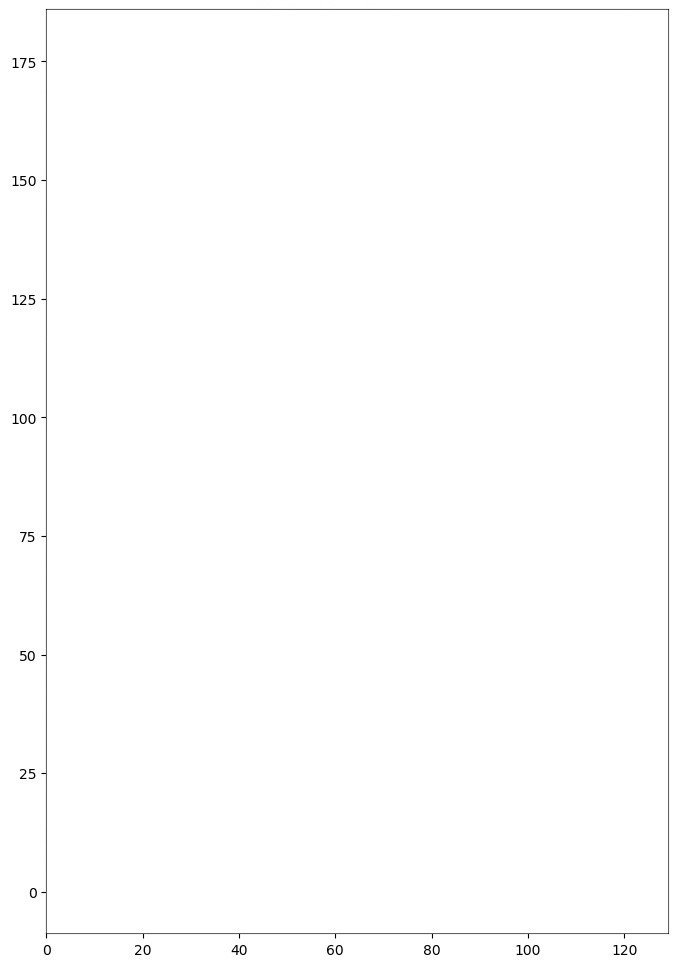}%
        \newline
        \includegraphics[width=0.11\linewidth]{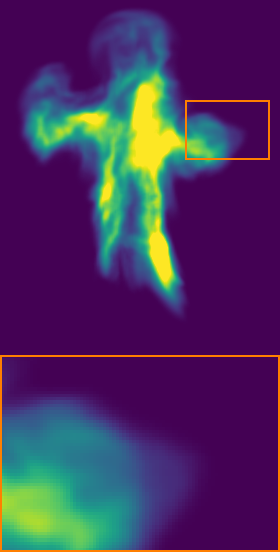}%
        \includegraphics[width=0.11\linewidth]{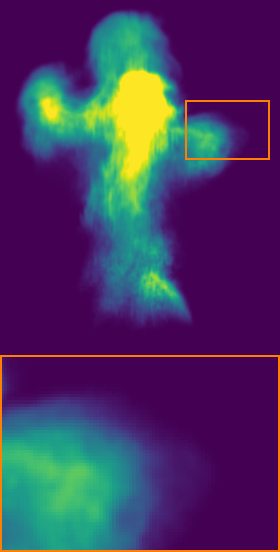}%
        \includegraphics[width=0.11\linewidth]{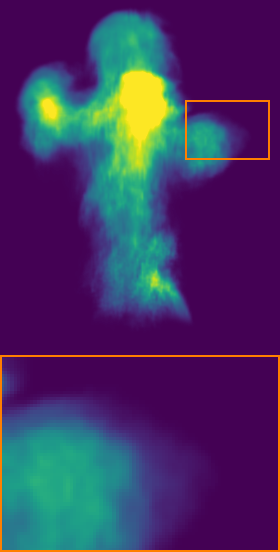}%
        \includegraphics[width=0.11\linewidth]{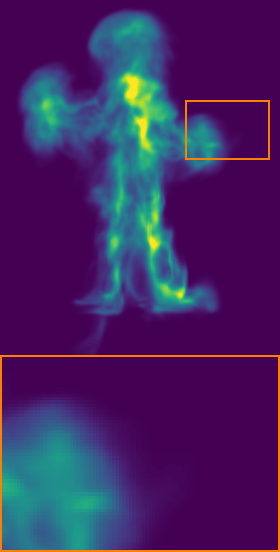}%
        \includegraphics[width=0.11\linewidth]{images/warp-vis_frame125/05_full_warp_final-side.png}%
        \includegraphics[width=0.11\linewidth]{images/warp-vis_frame125/29_targets_warp_final-side.png}%
        \hfill%
        \includegraphics[width=0.11\linewidth]{images/warp-vis_frame125/20_scalarFlow_warp_final-side.png}%
        \includegraphics[width=0.11\linewidth]{images/warp-vis_frame125/22_tomoFluid_warp_final-side.png}%
        \includegraphics[width=0.11\linewidth]{images/warp-vis_frame125/29_targets_warp_initial-side.png}%
    \vspace{-0.1cm}
    \caption{
    Multi-view evaluation with synthetic data: (a-e) Ablation with reference shown in (f). The different versions of the ablation continually improve density reconstruction (top) and motion accuracy, illustrated by advecting the initial state shown in (i) with the reconstructed velocity sequence, in the bottom row.
    (g-i) Comparison with previous work: 
    ScalarFlow \mycite{eckertScalarFlowLargescaleVolumetric2019}, TomoFluid \mycite{zangTomoFluidReconstructingDynamic2020}, and NeuralVolumes \mycite{lombardiNeuralVolumesLearning2019}. Our method in (e) yields an improved density reconstruction, in addition to a coherent, and physical transport. Note that (i) does not produce velocities.
    }
    \vspace{-0.2cm}
    \label{fig:eval2.synth}
\end{figure*}

\section{Additional Evaluation}\label{app:eval}
%
The Reference has a grid size of $128 \times 196 \times 128$.
As base grid size  for our reconstruction we use $128 \times 227 \times 128$,
which is cropped to the union of the AABBs of the visual hulls of all frames plus a padding of 4 cells on all sides.
For our reconstructions the grid sizes are therefore reduced to
$125 \times 187 \times 93$,
$94 \times 186 \times 95$,
$97 \times 164 \times 77$ and
$86 \times 161 \times 76$
for synthetic multi-view, synthetic single-view, real multi-view and real single-view for the base plume, respectively.
The spatial resolution remains the same.
As lighting model we use a single white point light with single-scattering, hand-placed above the central camera, and white ambient light to approximate multi-scattering.
The light intensities are $i_p=0.85$ and $i_a=0.64$, respectively.

\subsection{Additional Multi-View Ablations}
An extended evaluation of the motion of the synthetic case is shown in \myreffig{eval2.synth}.
The \emph{forward} reconstruction removes the streak-like artifacts, but does so at the expense of spatial reconstruction accuracy regarding the metrics. 
The step to \emph{coupled} improves the motion it results in only minor improvements in $\dens$.
The multi-scale approach (\varCoupledMS) brings the
metrics for $\dens$ on-par with the \textit{single} version.
The bottom row show a continuously improving transport of a test density, compared to the reference (\myreffig{eval2.synth} f), seen in the resulting distribution of the density.
While ScalarFlow might appear closer to the reference it shows a mismatch in shape at the stem and backside (right side, shown in the detail) of the resulting plume.

An ablation of multi-view reconstructions of the SF data is shown in \myreffig{eval.real.5view}.
Here, \varCoupledMS\ shows an unphysical accumulations of density at the top which are resolved by our \emph{global transport} (\varGlobWarp). The discriminator (\varDisc) has only little effect in the multi-view reconstruction, sometimes causing slight halo-artifacts, and is thus not used here. Previous methods give a more diffuse result.

\ef{
\subsection{Warp Loss in Global Transport} \label{app:eval.warploss}
In \myreffig{eval.denswarp} we show an additional ablation that highlights the effects of keeping the per-frame warp loss $\Ldwarp$ for both density ($\pderInl{\Ldwarp}{\dens}$) and velocity ($\pderInl{\Ldwarp}{\vel}$), even when global transport \myrefsecshort{method.transport.full} is used. Our global transport already provides gradients from the advection of the density for both $\dens$ \myrefeqshort{method.transport.ema} and $\vel$, meaning that the explicit density warping loss seems redundant for both.
In fact, since the sequence is defined via transport \myrefeqshort{method.transport.warp}, the warp loss, and therefore its gradients, should be zero. However, since the visual hull is applied after every advection step (in the initial transport, not the warp loss), the error stems from the parts removed by the hull. Using the hull as a loss, instead or in addition to a hard constraint, does not change anything about the results in \myreffig{eval.denswarp}.

As visible in the figure, $\pderInl{\Ldwarp}{\dens}$ is still mostly redundant, only in the real data case it reduces the divergence of the reconstructed density grid (not visible in the rendering). However, it also has no negative influence on the reconstruction.
Removing $\pderInl{\Ldwarp}{\vel}$ from the optimization has adverse effect in the real case, resulting in visible divergence in the lower part of the plume, turning the inflow into an outflow. In the Synthetic case the reconstruction still works well, showing only less inflow.
We conjecture that the differences between the synthetic and real cases are caused by a mismatch of the physical model to the real case.
Using $\Ldwarp$ puts more emphasis (gradients) on {\em local} correctness of the transport itself.
$\pderInl{\Ldwarp}{\dens}$ adapts the density to fit the existing transport. However, because the density also has to fit the observations, which have more weight, there is no change, and thus removing these gradients has no big impact.
$\pderInl{\Ldwarp}{\vel}$, on the other hand, adapts the velocity, \ie, the transport, to fit the existing density, including the hull constraint.
These results seem to suggest that explicit adaption of the local transport to the densities is still necessary when using the global transport.
}

\begin{figure*}[t!]%
    \centering%
    \begin{subfigure}[b]{0.57\linewidth}%
        \centering%
        \begin{tabular}{c|c|c|}%
             & w/ $\pderInl{\Ldenswarp}{\dens}$ & w/o $\pderInl{\Ldenswarp}{\dens}$ \\\hline%
            \shortstack{w/\\$\pder{\Ldenswarp}{\vel}$} & \includegraphics[width=0.34024\textwidth]{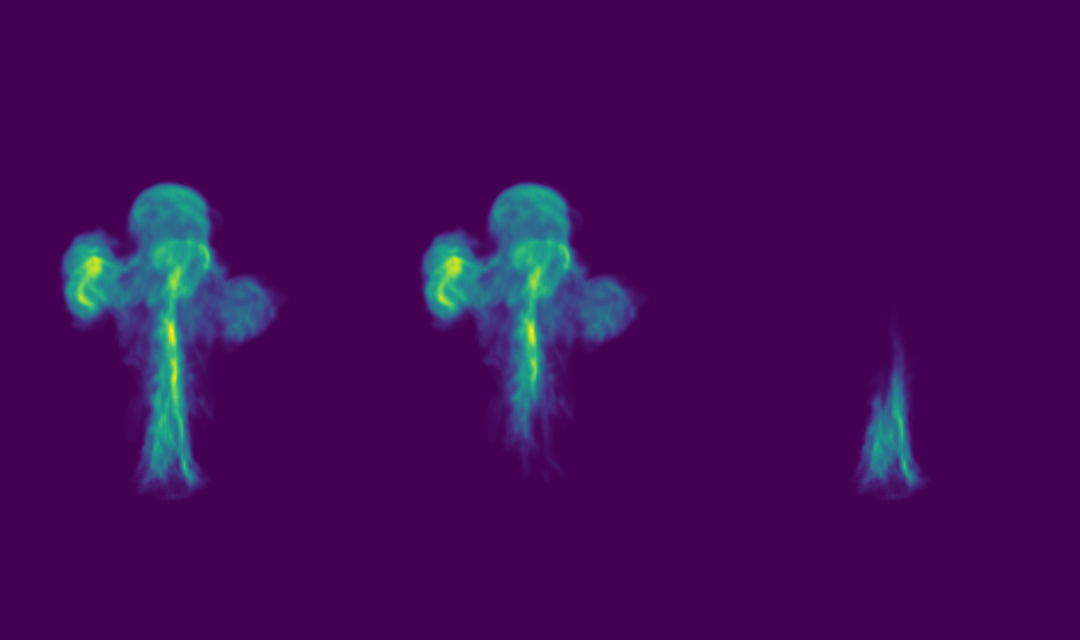} & \includegraphics[width=0.34024\textwidth]{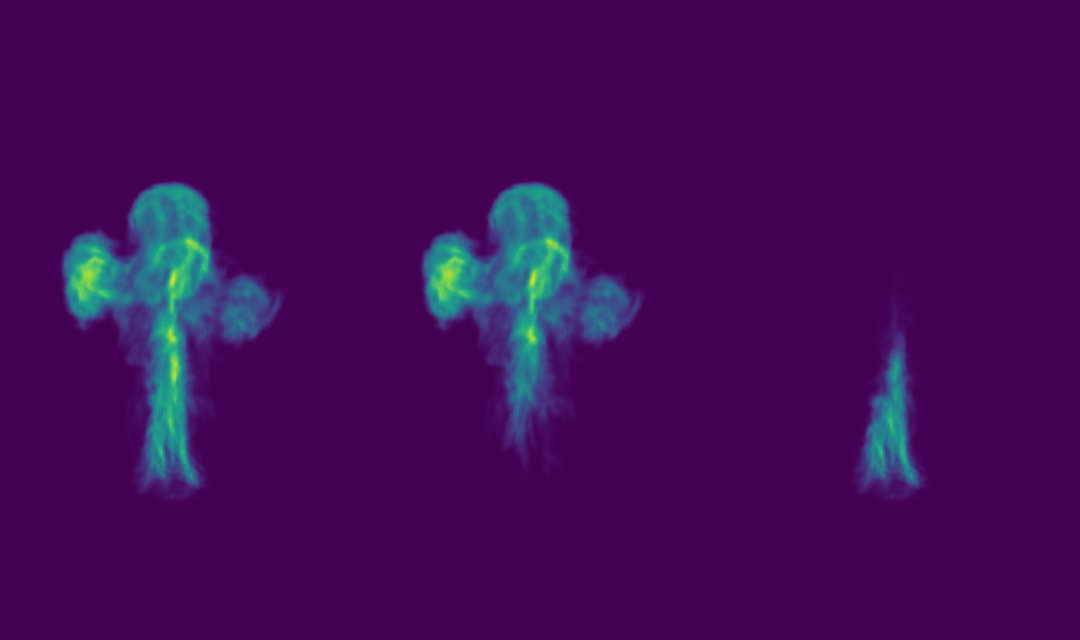} \\\hline%
            \shortstack{w/o\\$\pder{\Ldenswarp}{\vel}$} & \includegraphics[width=0.34024\textwidth]{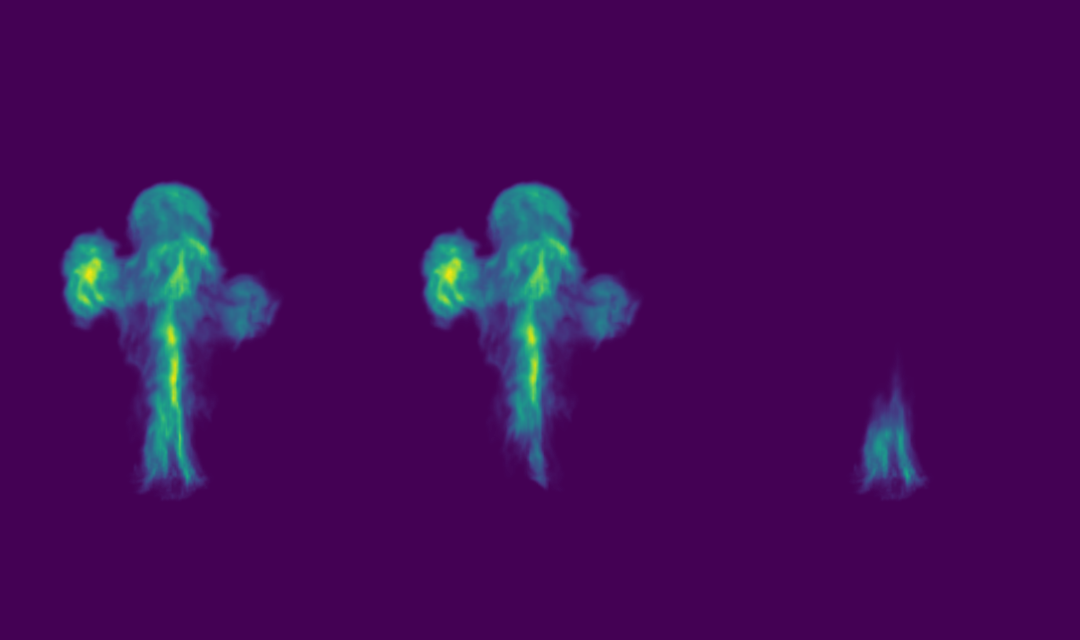} & \includegraphics[width=0.34024\textwidth]{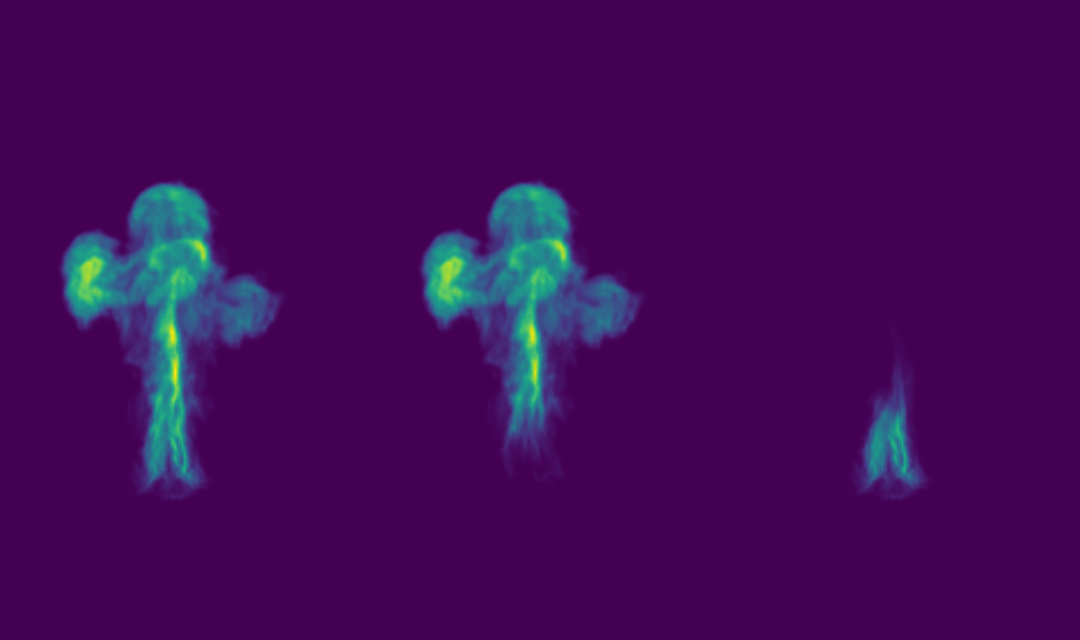} \\\hline%
        \end{tabular}%
        \caption{Synthetic}%
    \end{subfigure}%
    \begin{subfigure}[b]{0.43\linewidth}%
        \centering%
        \begin{tabular}{|c|c|}%
            w/ $\pderInl{\Ldenswarp}{\dens}$ & w/o $\pderInl{\Ldenswarp}{\dens}$ \\\hline%
            \includegraphics[width=0.45\textwidth]{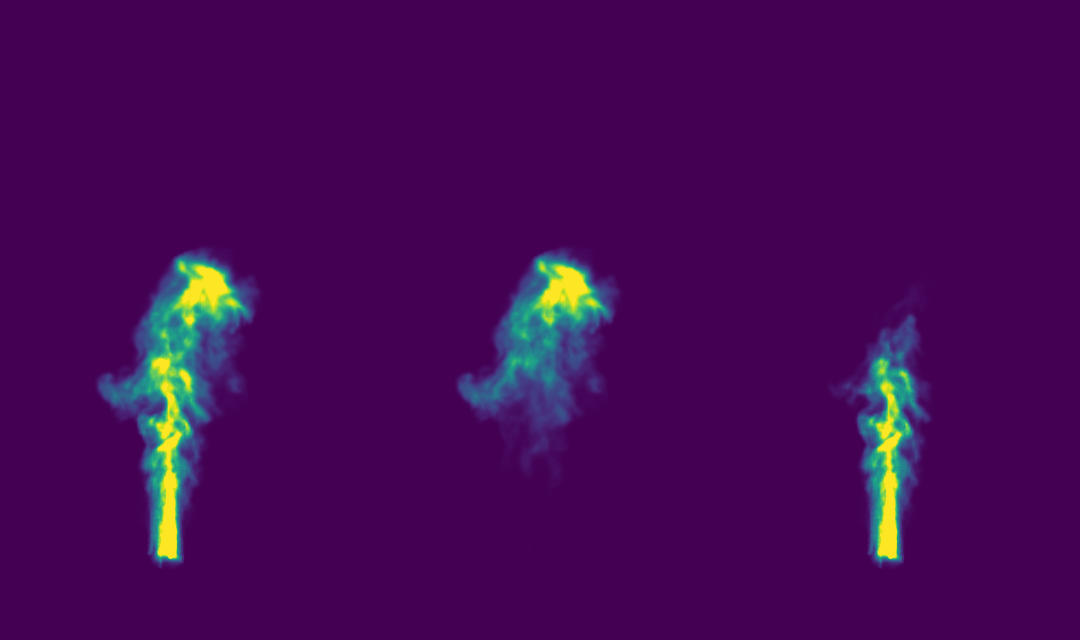} & \includegraphics[width=0.45\textwidth]{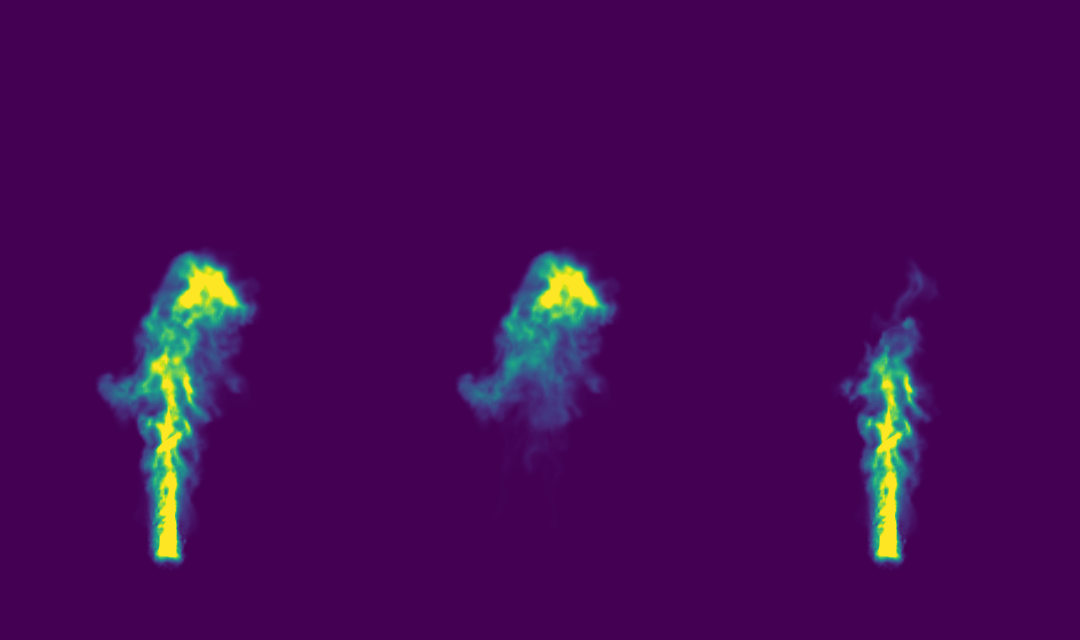} \\\hline%
            \includegraphics[width=0.45\textwidth]{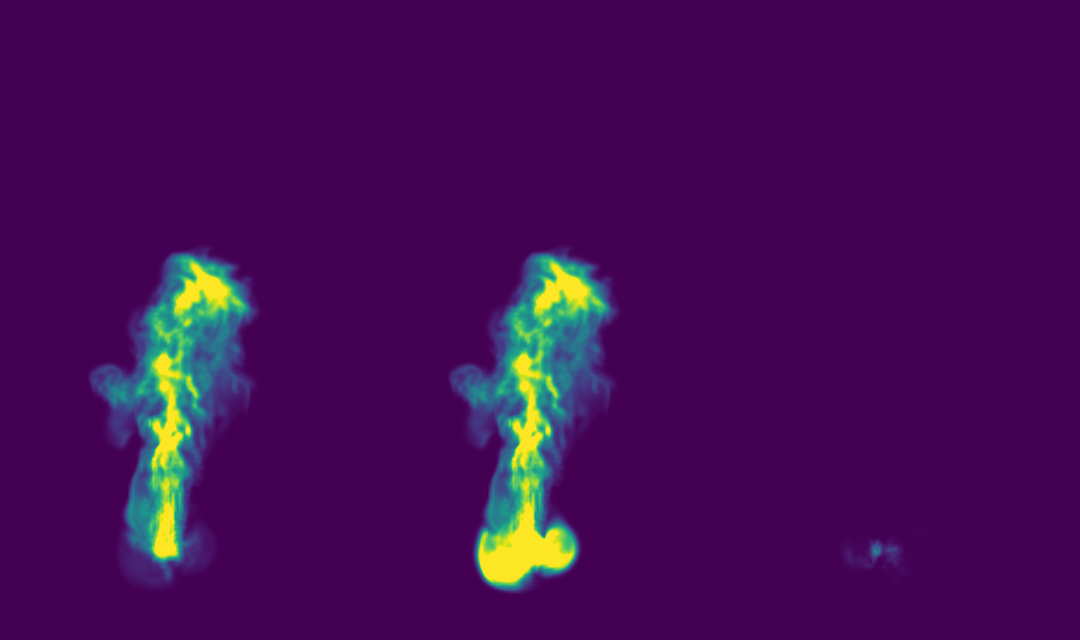} & \includegraphics[width=0.45\textwidth]{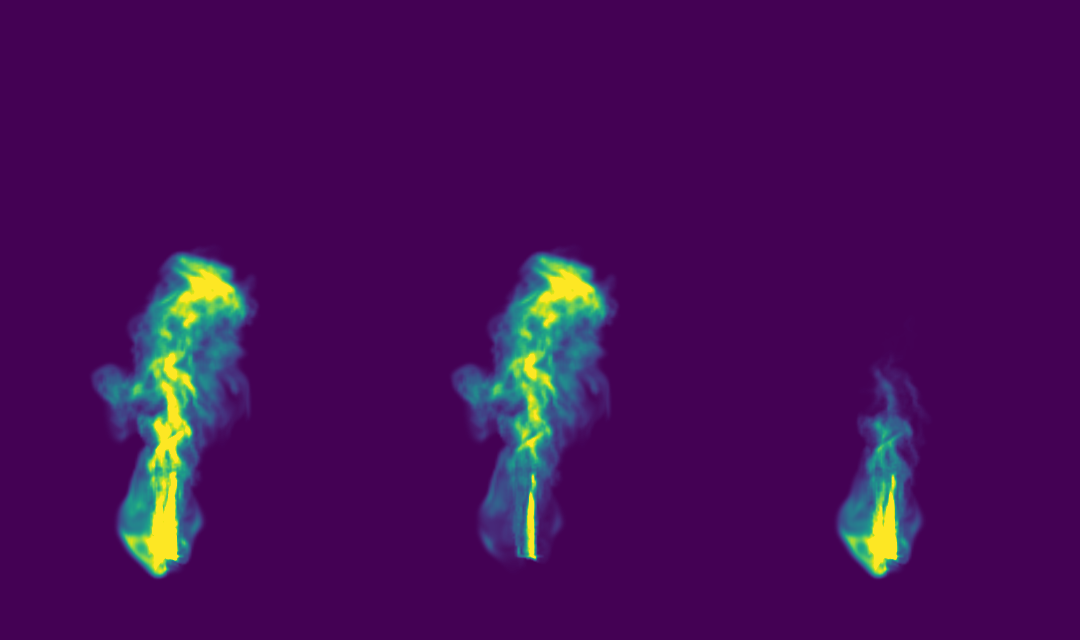} \\\hline%
        \end{tabular}%
        \caption{Real}%
    \end{subfigure}%
    \caption{\ef{Evaluation of the influence of the density warp loss $\Ldenswarp$ when {\em global transport} is also used.
    The tables show an ablation of the $\Ldenswarp$ applied to density and velocity, for both synthetic and real data.
    Every image shows the following triplet: left, the final result of the optimization, obtained by advecting $\dfirst$ with the velocity sequence and adding the inflow in every step; the center shows the same, but without any inflow; on the right, only the inflow is advected, \ie, $\dfirst$ is set to all zero.
    The top left images are the desired results.
    While the gradients provided by this loss are are almost redundant in the synthetic case (a), the gradients \wrt $\vel$ are necessary when using real data(b), even though the same dependencies are modeled by the global transport. Gradients \wrt $\dens$ are sill redundant, but do not hurt the reconstruction either.}}%
    \label{fig:eval.denswarp}%
\end{figure*}%

\subsection{State of the Art}
We re-implemented TomoFluid (TF) in our own Framework as no original source-code is publicly available as of the time of writing.
We use the loss functions and view-interpolations as described in the original work, but the hyperparameters had to be adapted to work in our setting.
As TF is implemented in our framework its grid is also cropped.

We use the original ScalarFlow code, which is based on mantaflow, the solver we used to create our synthetic test case. Due to this shared code-base SF has a certain advantage in the synthetic case. It shows in the visual similarity of the reconstructed motions compared to the reference, as seen in the stream-plots of \myreffig{eval.real.5view} (f,g).
Nevertheless, our method matches the targets better than the SF version.
The high transport error of SF in \myreftab{eval.synth} might be caused, in part, by the missing inflow as the ScalarFlow algorithm removes the synthetic inflow after reconstruction.
SF runs on the full $128 \times 196 \times 128$ grid in the synthetic test case while the real reconstructions from the provided dataset use a grid size of $100 \times 178 \times 100$.

NeuralVolumes typically works with many more viewpoints than we provide. This results in artifacts in our sparse-view setting, \ie, colored or black background is reconstructed within the volume.
The reconstruction is however still very good.
To compute metrics and rendering we sample a RGB$\alpha$ volume at the resolution of the reference and extract a density via
$\dens_{NV} := 0.03[\alpha]_0^1 (\text{R}+\text{G}+\text{B})/3$,
which removes black opaque artifacts, thus giving NV an advantage in these comparisons.

\newcommand{\mywidthMVc}{0.199\linewidth}
\begin{figure}
        \centering%
        \mygraphicsT{width=0.199\linewidth}{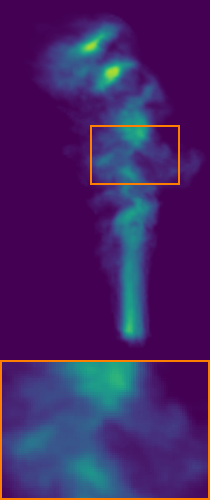}{a) \varCoupledMS}%
        \hfill%
        \mygraphicsT{width=0.199\linewidth}{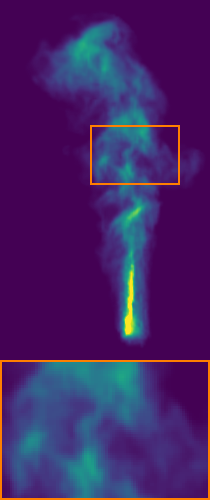}{b) \varGlobWarp}%
        \hfill%
        \mygraphicsT{width=0.199\linewidth}{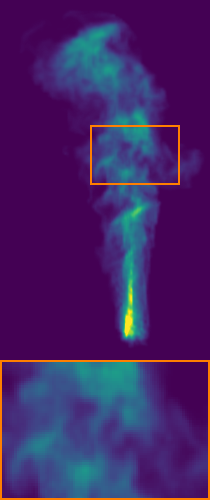}%
        {c) \varDisc}%
        \hfill%
        \mygraphicsT{width=0.199\linewidth}{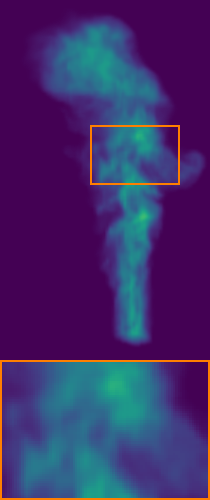}{d) SF \mycite{eckertScalarFlowLargescaleVolumetric2019}}%
        \hfill%
        \mygraphicsT{width=0.199\linewidth}{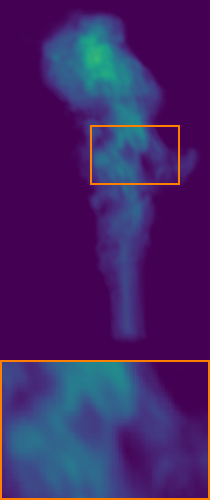}{e) NV \mycite{lombardiNeuralVolumesLearning2019}}%
        \newline%
        \includegraphics[width=0.199\linewidth]{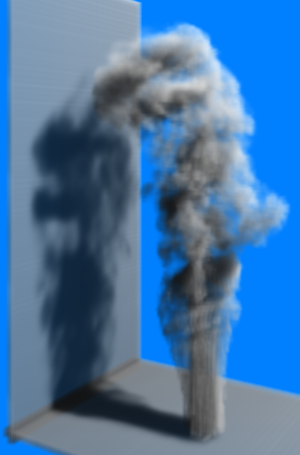}%
        \hfill%
        \includegraphics[width=0.199\linewidth]{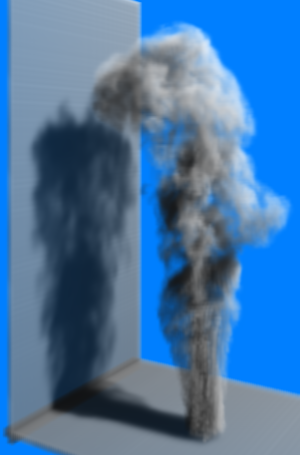}%
        \hfill%
        \includegraphics[width=0.199\linewidth]{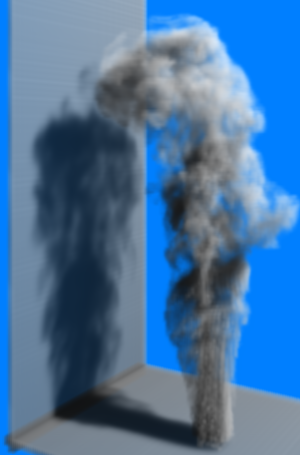}%
        \hfill%
        \includegraphics[width=0.199\linewidth]{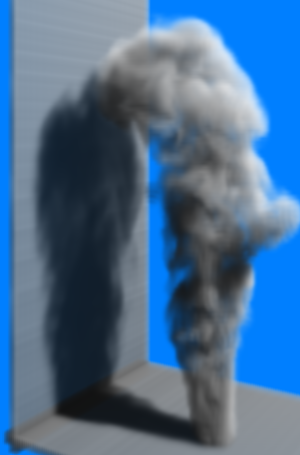}%
        \hfill%
        \includegraphics[width=0.199\linewidth]{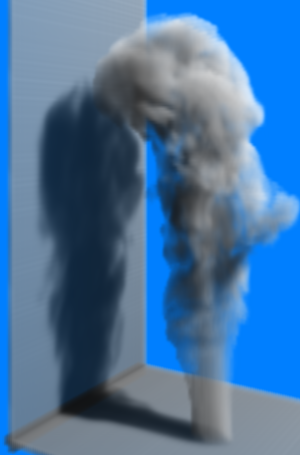}%
    \vspace{-0.1cm}
    \caption{
    Multi-views (5 targets) reconstruction with data from \mycite{eckertScalarFlowLargescaleVolumetric2019}.
    Top: perspective density from a $90^{\circ}$ view. 
    Bottom: rendering with lighting and inc. density from $60^{\circ}$.
    Global transport prevents artifacts, \eg, visible in (a),
    while previous work (d,e) is smoother.
    }
    \vspace{-0.2cm}
    \label{fig:eval.real.5view}
\end{figure}

\subsection{Details for the Single-View Ablation}
For the single-view ablation, \myreffig{eval.real.1view}, we focus on the key versions of our previous, multi-view ablation and evaluate them on real targets as the effect of the discriminator is more prominent here.
The tomographic reconstruction (a) yields a strong smearing out of information along the unconstrained direction of the viewing rays, despite the existence of transport initialization.
Although the qualitative examples from a $90^{\circ}$ angle in \myreffig{eval.real.1view}~top illustrate this behavior, it is even more clearly visible in the supplemental videos.
The \textit{coupling} version (b) refines the reconstructed volumes, but still contains noticeable striping artifacts as well as a lack of details.
Our \textit{global transport} formulation (c) resolves these artifacts and yields a plausible motion, which, however, does not adhere to the motions observed in the input views. 
In particular, it yields relatively strong single streaks of transported density
which are more diffusive in the real world flow.
The discriminator matches the appearance of the reconstruction to the real world smoke, thus guiding the motion reconstruction to adhere to features of real flows (d).

\clearpage
{\small
\bibliographystyle{ieee_fullname}
\bibliography{flowRecon,references-scalarflow}

\begin{thebibliography}{10}\itemsep=-1pt

\bibitem{Atcheson:2008:OEF}
Bradley Atcheson, Wolfgang Heidrich, and Ivo Ihrke.
\newblock An evaluation of optical flow algorithms for background oriented
  schlieren imaging.
\newblock {\em Experiments in Fluids}, 46(3):467--476, Mar 2009.

\bibitem{Atcheson2008}
Bradley Atcheson, Ivo Ihrke, Wolfgang Heidrich, Art Tevs, Derek Bradley, Marcus
  Magnor, and Hans-Peter Seidel.
\newblock Time-resolved 3d capture of non-stationary gas flows.
\newblock {\em ACM Trans. Graph.}, 27(5), Dec. 2008.

\bibitem{belden2010three}
Jesse Belden, Tadd~T Truscott, Michael~C Axiak, and Alexandra~H Techet.
\newblock Three-dimensional synthetic aperture particle image velocimetry.
\newblock {\em Measurement Science and Technology}, 21(12):125403, 2010.

\bibitem{chen2012thoracic}
Yang Chen, Zhou Yang, Yining Hu, Guanyu Yang, Yongcheng Zhu, Yinsheng Li, Wufan
  Chen, Christine Toumoulin, et~al.
\newblock Thoracic low-dose ct image processing using an artifact suppressed
  large-scale nonlocal means.
\newblock {\em Physics in Medicine \& Biology}, 57(9):2667, 2012.

\bibitem{blender}
Blender~Online Community.
\newblock {\em Blender - a 3D modelling and rendering package}.
\newblock Blender Foundation, Blender Institute, Amsterdam, 2020.

\bibitem{Dalziel00:BOS}
S.B. Dalziel, G.O. Hughes, and B.R. Sutherland.
\newblock {Whole-Field Density Measurements by 'Synthetic Schlieren'}.
\newblock {\em {Experiments in Fluids}}, 28:322--335, 2000.

\bibitem{de2018end}
Filipe de Avila Belbute-Peres, Kevin Smith, Kelsey Allen, Josh Tenenbaum, and
  J~Zico Kolter.
\newblock End-to-end differentiable physics for learning and control.
\newblock In {\em Advances in neural information processing systems}, 2018.

\bibitem{eckertCoupledFluidDensity2018}
M.-L. Eckert, W. Heidrich, and N. Thuerey.
\newblock Coupled {{Fluid Density}} and {{Motion}} from {{Single Views}}.
\newblock {\em Computer Graphics Forum}, 37(8):47--58, 2018.

\bibitem{eckertScalarFlowLargescaleVolumetric2019}
Marie-Lena Eckert, Kiwon Um, and Nils Thuerey.
\newblock {{ScalarFlow}}: A large-scale volumetric data set of real-world
  scalar transport flows for computer animation and machine learning.
\newblock {\em ACM Transactions on Graphics (TOG)}, 38(6):1--16, 2019.

\bibitem{efros2001image}
Alexei~A Efros and William~T Freeman.
\newblock Image quilting for texture synthesis and transfer.
\newblock In {\em Proceedings of the 28th annual conference on Computer
  graphics and interactive techniques}, pages 341--346, 2001.

\bibitem{elsinga2006tomographic}
Gerrit~E Elsinga, Fulvio Scarano, Bernhard Wieneke, and Bas~W van Oudheusden.
\newblock Tomographic particle image velocimetry.
\newblock {\em Experiments in fluids}, 41(6):933--947, 2006.

\bibitem{Elsinga:06}
Gerrit~E Elsinga, Fulvio Scarano, Bernhard Wieneke, and Bas~W van Oudheusden.
\newblock Tomographic particle image velocimetry.
\newblock {\em Experiments in fluids}, 41(6):933--947, 2006.

\bibitem{engel2004real}
Klaus Engel, Markus Hadwiger, Joe~M Kniss, Aaron~E Lefohn, Christof~Rezk
  Salama, and Daniel Weiskopf.
\newblock Real-time volume graphics.
\newblock In {\em ACM Siggraph 2004 Course Notes}, pages 29--es. 2004.

\bibitem{fahringer2015volumetric}
Timothy~W Fahringer, Kyle~P Lynch, and Brian~S Thurow.
\newblock Volumetric particle image velocimetry with a single plenoptic camera.
\newblock {\em Measurement Science and Technology}, 26(11):115201, 2015.

\bibitem{fedkiw:2001:VSO}
Ronald Fedkiw, Jos Stam, and Henrik~Wann Jensen.
\newblock Visual simulation of smoke.
\newblock In Eugene Fiume, editor, {\em Proceedings of SIGGRAPH 2001}, Computer
  Graphics Proceedings, Annual Conference Series, pages 15--22. ACM, ACM Press
  / ACM SIGGRAPH, 2001.

\bibitem{fuchs2007density}
Christian Fuchs, Tongbo Chen, Michael Goesele, Holger Theisel, and Hans-Peter
  Seidel.
\newblock Density estimation for dynamic volumes.
\newblock {\em Computers \& Graphics}, 31(2):205--211, 2007.

\bibitem{Gkioulekas2016}
Ioannis Gkioulekas, Anat Levin, and Todd Zickler.
\newblock An evaluation of computational imaging techniques for heterogeneous
  inverse scattering.
\newblock In Bastian Leibe, Jiri Matas, Nicu Sebe, and Max Welling, editors,
  {\em Computer Vision -- ECCV 2016}, pages 685--701, Cham, 2016. Springer
  International Publishing.

\bibitem{Goldhahn07}
E. Goldhahn and J. Seume.
\newblock {The Background Oriented Schlieren Technique: Sensitivity, Accuracy,
  Resolution and Application to a Three-Dimensional Density Field}.
\newblock {\em Experiments in Fluids}, 43(2--3):241--249, 2007.

\bibitem{goodfellow2014generative}
Ian~J Goodfellow, Jean Pouget-Abadie, Mehdi Mirza, Bing Xu, David Warde-Farley,
  Sherjil Ozair, Aaron Courville, and Yoshua Bengio.
\newblock Generative adversarial nets.
\newblock {\em stat}, 1050:10, 2014.

\bibitem{Grant97:PIV}
I. Grant.
\newblock {Particle Image Velocimetry: a Review}.
\newblock {\em Proceedings of the Institution of Mechanical Engineers},
  211(1):55--76, 1997.

\bibitem{Gregson:2014}
James Gregson, Ivo Ihrke, Nils Thuerey, and Wolfgang Heidrich.
\newblock From capture to simulation: connecting forward and inverse problems
  in fluids.
\newblock {\em {ACM} Trans. Graph.}, 33(4):139, 2014.

\bibitem{Gregson:2012}
James Gregson, Michael Krimerman, Matthias~B Hullin, and Wolfgang Heidrich.
\newblock Stochastic tomography and its applications in 3d imaging of mixing
  fluids.
\newblock {\em {ACM} Trans. Graph.}, 31(4):52--1, 2012.

\bibitem{Gu:13}
Jinwei Gu, S.K. Nayar, E. Grinspun, P.N. Belhumeur, and R. Ramamoorthi.
\newblock {Compressive Structured Light for Recovering Inhomogeneous
  Participating Media}.
\newblock {\em {IEEE} Trans. Pattern Analysis and Mach. Int.}, 35(3):555--567,
  2013.

\bibitem{Hawkins:timeVary}
Tim Hawkins, Per Einarsson, and Paul Debevec.
\newblock Acquisition of time-varying participating media.
\newblock {\em {ACM} Trans. Graph.}, 24(3):812--815, 2005.

\bibitem{henzlerEscapingPlatoCave2019}
Philipp Henzler, Niloy~J. Mitra, and Tobias Ritschel.
\newblock Escaping {{Plato}}'s {{Cave}}: {{3D Shape From Adversarial
  Rendering}}.
\newblock In {\em Proceedings of the {{IEEE International Conference}} on
  {{Computer Vision}}}, pages 9984--9993, 2019.

\bibitem{henzlerSingleimageTomography3D2018}
Phlipp Henzler, Volker Rasche, Timo Ropinski, and Tobias Ritschel.
\newblock Single-image {{Tomography}}: {{3D Volumes}} from {{2D Cranial
  X}}-{{Rays}}.
\newblock {\em Computer Graphics Forum}, 37(2):377--388, 2018.

\bibitem{heusel2017gans}
Martin Heusel, Hubert Ramsauer, Thomas Unterthiner, Bernhard Nessler, and Sepp
  Hochreiter.
\newblock Gans trained by a two time-scale update rule converge to a local nash
  equilibrium.
\newblock In {\em Advances in neural information processing systems}, pages
  6626--6637, 2017.

\bibitem{hollLearningControlPDEs2019}
Philipp Holl, Nils Thuerey, and Vladlen Koltun.
\newblock Learning to {{Control PDEs}} with {{Differentiable Physics}}.
\newblock In {\em International {{Conference}} on {{Learning
  Representations}}}, 2020.

\bibitem{Hu2020}
Yuanming Hu, Luke Anderson, Tzu-Mao Li, Qi Sun, Nathan Carr, Jonathan
  Ragan-Kelley, and Fredo Durand.
\newblock Differentiable programming for physical simulation.
\newblock In {\em International Conference on Learning Representations}, 2020.

\bibitem{Ihrke:tomoFlames}
Ivo Ihrke and Marcus Magnor.
\newblock Image-based tomographic reconstruction of flames.
\newblock In {\em ACM SIGGRAPH/Eurographics Symposium on Computer Animation},
  pages 365--373. Eurographics Association, 2004.

\bibitem{islamHowMuchPosition2020}
Amirul Islam, Sen Jia, and D.~B.~Neil Bruce.
\newblock How much position information do convolutional neural networks
  encode.
\newblock {\em International Conference on Learning Representations}, 2020.

\bibitem{javed2005appearance}
Omar Javed, Khurram Shafique, and Mubarak Shah.
\newblock Appearance modeling for tracking in multiple non-overlapping cameras.
\newblock In {\em 2005 IEEE Computer Society Conference on Computer Vision and
  Pattern Recognition (CVPR'05)}, volume~2, pages 26--33. IEEE, 2005.

\bibitem{Ji_2013_CVPR}
Yu Ji, Jinwei Ye, and Jingyi Yu.
\newblock Reconstructing gas flows using light-path approximation.
\newblock In {\em Proceedings of the IEEE Conference on Computer Vision and
  Pattern Recognition (CVPR)}, June 2013.

\bibitem{jolicoeur-martineauRelativisticDiscriminatorKey2018a}
Alexia Jolicoeur-Martineau.
\newblock The relativistic discriminator: A key element missing from standard
  {{GAN}}.
\newblock In {\em International {{Conference}} on {{Learning
  Representations}}}, 2018.

\bibitem{kanazawaLearningCategorySpecificMesh2018}
Angjoo Kanazawa, Shubham Tulsiani, Alexei~A. Efros, and Jitendra Malik.
\newblock Learning {{Category}}-{{Specific Mesh Reconstruction}} from {{Image
  Collections}}.
\newblock In {\em Proceedings of the {{European Conference}} on {{Computer
  Vision}} ({{ECCV}})}, pages 371--386, 2018.

\bibitem{karras2019style}
Tero Karras, Samuli Laine, and Timo Aila.
\newblock A style-based generator architecture for generative adversarial
  networks.
\newblock In {\em Proceedings of the IEEE conference on computer vision and
  pattern recognition}, pages 4401--4410, 2019.

\bibitem{katoNeural3DMesh2018}
Hiroharu Kato, Yoshitaka Ushiku, and Tatsuya Harada.
\newblock Neural {{3D Mesh Renderer}}.
\newblock In {\em Proceedings of the {{IEEE Conference}} on {{Computer Vision}}
  and {{Pattern Recognition}}}, pages 3907--3916, 2018.

\bibitem{kato2018neural}
Hiroharu Kato, Yoshitaka Ushiku, and Tatsuya Harada.
\newblock Neural 3d mesh renderer.
\newblock In {\em IEEE Conference on Computer Vision and Pattern Recognition
  (CVPR)}, 2018.

\bibitem{Kim2019}
Byungsoo Kim, Vinicius~C. Azevedo, Markus Gross, and Barbara Solenthaler.
\newblock {Transport-based neural style transfer for smoke simulations}.
\newblock {\em ACM Transactions on Graphics}, 38(6):1--11, nov 2019.

\bibitem{Kim2020}
Byungsoo Kim, Vinicius~C. Azevedo, Markus Gross, and Barbara Solenthaler.
\newblock Lagrangian neural style transfer for fluids.
\newblock {\em ACM Transactions on Graphics}, 39(4), July 2020.

\bibitem{kingma2014adam}
Diederik Kingma and Jimmy Ba.
\newblock Adam: A method for stochastic optimization.
\newblock {\em International Conference on Learning Representations (ICLR)},
  2015.

\bibitem{koutsoudis2014multi}
Anestis Koutsoudis, Bla{\v{z}} Vidmar, George Ioannakis, Fotis Arnaoutoglou,
  George Pavlidis, and Christodoulos Chamzas.
\newblock Multi-image 3d reconstruction data evaluation.
\newblock {\em Journal of Cultural Heritage}, 15(1):73--79, 2014.

\bibitem{Levis_2015_ICCV}
Aviad Levis, Yoav~Y. Schechner, Amit Aides, and Anthony~B. Davis.
\newblock Airborne three-dimensional cloud tomography.
\newblock In {\em Proceedings of the IEEE International Conference on Computer
  Vision (ICCV)}, December 2015.

\bibitem{Li2018DMC}
Tzu-Mao Li, Miika Aittala, Fr{\'e}do Durand, and Jaakko Lehtinen.
\newblock Differentiable monte carlo ray tracing through edge sampling.
\newblock {\em ACM Trans. Graph. (Proc. SIGGRAPH Asia)}, 37(6):222:1--222:11,
  2018.

\bibitem{lombardiNeuralVolumesLearning2019}
Stephen Lombardi, Tomas Simon, Jason Saragih, Gabriel Schwartz, Andreas
  Lehrmann, and Yaser Sheikh.
\newblock Neural {{Volumes}}: {{Learning Dynamic Renderable Volumes}} from
  {{Images}}.
\newblock {\em ACM Transactions on Graphics}, 38(4):1--14, 2019.

\bibitem{Loper2014}
Matthew~M. Loper and Michael~J. Black.
\newblock {OpenDR}: An approximate differentiable renderer.
\newblock In {\em Computer Vision -- ECCV 2014}, volume 8695 of {\em Lecture
  Notes in Computer Science}, pages 154--169, Sept. 2014.

\bibitem{mantaflow}
{MantaFlow}, 2018.
\newblock {http://mantaflow.com}.

\bibitem{McNamara2004}
Antoine McNamara, Adrien Treuille, Zoran Popovi\'{c}, and Jos Stam.
\newblock Fluid control using the adjoint method.
\newblock {\em ACM Trans. Graph.}, 23(3):449–456, Aug. 2004.

\bibitem{meinhardt2013horn}
Enric Meinhardt-Llopis, Javier~S{\'a}nchez P{\'e}rez, and Daniel Kondermann.
\newblock Horn-schunck optical flow with a multi-scale strategy.
\newblock {\em Image Processing on line}, 2013:151--172, 2013.

\bibitem{mescheder2019occnet}
Lars Mescheder, Michael Oechsle, Michael Niemeyer, Sebastian Nowozin, and
  Andreas Geiger.
\newblock Occupancy networks: Learning 3d reconstruction in function space.
\newblock In {\em Proceedings of the IEEE Conference on Computer Vision and
  Pattern Recognition}, pages 4460--4470, 2019.

\bibitem{mildenhall2020nerf}
Ben Mildenhall, Pratul~P. Srinivasan, Matthew Tancik, Jonathan~T. Barron, Ravi
  Ramamoorthi, and Ren Ng.
\newblock Nerf: Representing scenes as neural radiance fields for view
  synthesis.
\newblock In {\em ECCV}, 2020.

\bibitem{moon2018v2v}
Gyeongsik Moon, Ju Yong~Chang, and Kyoung Mu~Lee.
\newblock V2v-posenet: Voxel-to-voxel prediction network for accurate 3d hand
  and human pose estimation from a single depth map.
\newblock In {\em Proceedings of the IEEE conference on computer vision and
  pattern Recognition}, pages 5079--5088, 2018.

\bibitem{morris2011dynamic}
Nigel~JW Morris and Kiriakos~N Kutulakos.
\newblock Dynamic refraction stereo.
\newblock {\em IEEE transactions on pattern analysis and machine intelligence},
  33(8):1518--1531, 2011.

\bibitem{musialski2013survey}
Przemyslaw Musialski, Peter Wonka, Daniel~G Aliaga, Michael Wimmer, Luc
  Van~Gool, and Werner Purgathofer.
\newblock A survey of urban reconstruction.
\newblock In {\em Computer graphics forum}, volume 32(6), pages 146--177. Wiley
  Online Library, 2013.

\bibitem{niemeyer2019occupancy}
Michael Niemeyer, Lars Mescheder, Michael Oechsle, and Andreas Geiger.
\newblock Occupancy flow: 4d reconstruction by learning particle dynamics.
\newblock In {\em Proceedings of the IEEE International Conference on Computer
  Vision}, pages 5379--5389, 2019.

\bibitem{Nimier2019Mitsuba2}
Merlin Nimier-David, Delio Vicini, Tizian Zeltner, and Wenzel Jakob.
\newblock Mitsuba 2: A retargetable forward and inverse renderer.
\newblock {\em Transactions on Graphics (Proceedings of SIGGRAPH Asia)}, 38(6),
  Dec. 2019.

\bibitem{okabeFluidVolumeModeling2015}
Makoto Okabe, Yoshinori Dobashi, Ken Anjyo, and Rikio Onai.
\newblock Fluid volume modeling from sparse multi-view images by appearance
  transfer.
\newblock {\em ACM Transactions on Graphics}, 34(4):1--10, 2015.

\bibitem{papon2013voxel}
Jeremie Papon, Alexey Abramov, Markus Schoeler, and Florentin Worgotter.
\newblock Voxel cloud connectivity segmentation-supervoxels for point clouds.
\newblock In {\em Proceedings of the IEEE conference on computer vision and
  pattern recognition}, pages 2027--2034, 2013.

\bibitem{peskin1972flow}
Charles~S Peskin.
\newblock Flow patterns around heart valves: a numerical method.
\newblock {\em Journal of computational physics}, 10(2):252--271, 1972.

\bibitem{pope2001turbulent}
Stephen~B Pope.
\newblock Turbulent flows, 2001.

\bibitem{qian2017stereo}
Yiming Qian, Minglun Gong, and Yee-Hong Yang.
\newblock Stereo-based 3d reconstruction of dynamic fluid surfaces by global
  optimization.
\newblock In {\em Proceedings of the IEEE conference on computer vision and
  pattern recognition}, pages 1269--1278, 2017.

\bibitem{RadfordMC15}
Alec Radford, Luke Metz, and Soumith Chintala.
\newblock Unsupervised representation learning with deep convolutional
  generative adversarial networks.
\newblock {\em Proc. ICLR}, 2016.

\bibitem{Schenk2018}
Connor Schenck and Dieter Fox.
\newblock Spnets: Differentiable fluid dynamics for deep neural networks.
\newblock In {\em Conference on Robot Learning}, volume~87 of {\em Proceedings
  of Machine Learning Research}, 2018.

\bibitem{schenck2018spnets}
Connor Schenck and Dieter Fox.
\newblock Spnets: Differentiable fluid dynamics for deep neural networks.
\newblock In {\em Conference on Robot Learning}, pages 317--335, 2018.

\bibitem{selleUnconditionallyStableMacCormack2008}
Andrew Selle, Ronald Fedkiw, ByungMoon Kim, Yingjie Liu, and Jarek Rossignac.
\newblock An {{Unconditionally Stable MacCormack Method}}.
\newblock {\em Journal of Scientific Computing}, 35(2):350--371, 2008.

\bibitem{shen2019r}
Tiancheng Shen, Xia Li, Zhisheng Zhong, Jianlong Wu, and Zhouchen Lin.
\newblock R-net: Recurrent and recursive network for sparse-view ct artifacts
  removal.
\newblock In {\em International Conference on Medical Image Computing and
  Computer-Assisted Intervention}, pages 319--327. Springer, 2019.

\bibitem{shur1999detached}
M Shur, PR Spalart, M Strelets, and A Travin.
\newblock Detached-eddy simulation of an airfoil at high angle of attack.
\newblock In {\em Engineering turbulence modelling and experiments 4}, pages
  669--678. Elsevier, 1999.

\bibitem{sidky2008image}
Emil~Y Sidky and Xiaochuan Pan.
\newblock Image reconstruction in circular cone-beam computed tomography by
  constrained, total-variation minimization.
\newblock {\em Physics in Medicine \& Biology}, 53(17):4777, 2008.

\bibitem{sitzmann2019deepvoxels}
Vincent Sitzmann, Justus Thies, Felix Heide, Matthias Nie{\ss}ner, Gordon
  Wetzstein, and Michael Zollhofer.
\newblock Deepvoxels: Learning persistent 3d feature embeddings.
\newblock In {\em Proceedings of the IEEE Conference on Computer Vision and
  Pattern Recognition}, pages 2437--2446, 2019.

\bibitem{sitzmann2019scene}
Vincent Sitzmann, Michael Zollh{\"o}fer, and Gordon Wetzstein.
\newblock Scene representation networks: Continuous 3d-structure-aware neural
  scene representations.
\newblock In {\em Advances in Neural Information Processing Systems}, pages
  1121--1132, 2019.

\bibitem{tewari2020state}
Ayush Tewari, Ohad Fried, Justus Thies, Vincent Sitzmann, Stephen Lombardi,
  Kalyan Sunkavalli, Ricardo Martin-Brualla, Tomas Simon, Jason Saragih,
  Matthias Nießner, Rohit Pandey, Sean Fanello, Gordon Wetzstein, Jun-Yan Zhu,
  Christian Theobalt, Maneesh Agrawala, Eli Shechtman, Dan~B Goldman, and
  Michael Zollhöfer.
\newblock State of the art on neural rendering, 2020.

\bibitem{thapa2020dynamic}
Simron Thapa, Nianyi Li, and Jinwei Ye.
\newblock Dynamic fluid surface reconstruction using deep neural network.
\newblock In {\em Proceedings of the IEEE/CVF Conference on Computer Vision and
  Pattern Recognition}, pages 21--30, 2020.

\bibitem{um2020sol}
Kiwon Um, Robert Brand, Philipp Holl, and Nils Fei, Raymond~Thuerey.
\newblock Solver-in-the-loop: Learning from differentiable physics to interact
  with iterative pde-solvers.
\newblock {\em Advances in Neural Information Processing Systems}, 2020.

\bibitem{wang2004ssim}
Zhou Wang, Alan~C Bovik, Hamid~R Sheikh, and Eero~P Simoncelli.
\newblock Image quality assessment: from error visibility to structural
  similarity.
\newblock {\em IEEE transactions on image processing}, 13(4):600--612, 2004.

\bibitem{weiss2020ssc}
Sebastian Weiss, Robert Maier, Daniel Cremers, Rudiger Westermann, and Nils
  Thuerey.
\newblock Correspondence-free material reconstruction using sparse surface
  constraints.
\newblock pages 4686--4695, 2020.

\bibitem{xieTempoGANTemporallyCoherent2018}
You Xie, Erik Franz, Mengyu Chu, and Nils Thuerey.
\newblock {{tempoGAN}}: A temporally coherent, volumetric {{GAN}} for
  super-resolution fluid flow.
\newblock {\em ACM Transactions on Graphics}, 37(4):95:1--95:15, 2018.

\bibitem{Xiong:2017:rainbowPIV}
Jinhui Xiong, Ramzi Idoughi, Andres~A Aguirre-Pablo, Abdulrahman~B Aljedaani,
  Xiong Dun, Qiang Fu, Sigurdur~T Thoroddsen, and Wolfgang Heidrich.
\newblock Rainbow particle imaging velocimetry for dense 3d fluid velocity
  imaging.
\newblock {\em {ACM} Trans. Graph.}, 36(4):36, 2017.

\bibitem{Zang2018}
Guangming Zang, Ramzi Idoughi, Ran Tao, Gilles Lubineau, Peter Wonka, and
  Wolfgang Heidrich.
\newblock Space-time tomography for continuously deforming objects.
\newblock {\em ACM Trans. Graph.}, 37(4), July 2018.

\bibitem{zangTomoFluidReconstructingDynamic2020}
Guangming Zang, Ramzi Idoughi, Congli Wang, Anthony Bennett, Jianguo Du, Scott
  Skeen, William~L. Roberts, Peter Wonka, and Wolfgang Heidrich.
\newblock {{TomoFluid}}: {{Reconstructing Dynamic Fluid}} from {{Sparse View
  Videos}}.
\newblock In {\em Proceedings of the {{IEEE Conference}} on {{Computer Vision}}
  and {{Pattern Recognition}}}. {IEEE}, 2020.

\bibitem{zhang2018lpips}
Richard Zhang, Phillip Isola, Alexei~A Efros, Eli Shechtman, and Oliver Wang.
\newblock The unreasonable effectiveness of deep features as a perceptual
  metric.
\newblock In {\em Proceedings of the IEEE conference on computer vision and
  pattern recognition}, pages 586--595, 2018.

\bibitem{zhu2017unpaired}
Jun-Yan Zhu, Taesung Park, Phillip Isola, and Alexei~A Efros.
\newblock Unpaired image-to-image translation using cycle-consistent
  adversarial networks.
\newblock In {\em Proceedings of the IEEE international conference on computer
  vision}, pages 2223--2232, 2017.

\end{thebibliography}
}

\end{document}